\documentclass{article}

\usepackage[nonatbib, preprint]{neurips_2024}
\usepackage[numbers]{natbib}
\usepackage{xcolor}         

\usepackage[utf8]{inputenc} 
\usepackage{mathtools} 

\usepackage{times}
\usepackage{soul}
\usepackage{amsmath}
\usepackage{colortbl} 
\usepackage{subcaption} 
\usepackage{amssymb}
\newcommand{\1}{\text{1\hspace{-1.5mm}1}} 
\usepackage{bm}
\usepackage{algorithm}
\usepackage{algorithmicx}
\usepackage{algpseudocode}
\usepackage{graphicx} 
\usepackage{hyperref}       
\usepackage{wrapfig}
\usepackage{tcolorbox}
\usepackage{placeins} 
\usepackage{mdframed}
\usepackage[capitalize,noabbrev]{cleveref}

\usepackage{booktabs} 
\usepackage{colortbl} 
\newtcolorbox{myquote}[1][]{colback=gray!10, colframe=black, fonttitle=\bfseries, title=#1, sharp corners}

\title{Revisiting SMoE Language Models by Evaluating Inefficiencies with Task Specific Expert Pruning}

\author{Soumajyoti Sarkar\thanks{Amazon AGI, work performed while at Amazon Web Services\ \{soumajs@amazon.com, zhasheng@amazon.com\}} 
\And 
        Leonard Lausen\thanks{Amazon Web Services \ \{lausen@amazon.com, gkarypis@amazon.com\}}%
\And 
        Volkan Cevher\thanks{Amazon AGI \{volkcevh@amazon.de\} and LIONS, EPFL. Volkan holds joint appointment at Amazon and EPFL. This paper represents the work performed at Amazon.}
\And
        Sheng Zha$^*$
\AND 
        Thomas Brox\thanks{Department of Computer Science, University of Freiburg, Germany. This paper represents the work performed at Amazon.}
\And    
        George Karypis$^\dagger$
}
\begin{document}

\maketitle

\begin{abstract}
Sparse Mixture of Expert (SMoE) models have emerged as a scalable alternative to dense models in language modeling. These models use conditionally activated feedforward subnetworks in transformer blocks, allowing for a separation between total model parameters and per-example computation. However, large token-routed SMoE models face a significant challenge: during inference, the entire model must be used for a sequence or a batch, resulting in high latencies in a distributed setting that offsets the advantages of per-token sparse activation.
Our research explores task-specific model pruning to inform decisions about designing SMoE architectures, mainly modulating the choice of expert counts in pretraining. We investigate whether such pruned models offer advantages over smaller SMoE models trained from scratch, when evaluating and comparing them individually on tasks. To that end, we introduce an adaptive task-aware pruning technique {\tt UNCURL} to reduce the number of experts per MoE layer in an offline manner post-training. 
Our findings reveal a threshold pruning factor for the reduction that depends on the number of experts used in pretraining, above which, the reduction starts to degrade model performance. These insights contribute to our understanding of model design choices when pretraining with SMoE architectures, particularly useful when considering task-specific inference optimization for later stages.



\end{abstract}

\section{Introduction}


Conditional computation in language models, where parts of the network are active on a per-example basis, has been proposed in theory as a way to dramatically increase model capacity without a proportional increase in computation needed per token. In most of these schemes, large parts of the network are inactive on a per-token basis \citep{jacobs1991adaptive,shazeer2017outrageously}. Formally, an SMoE model \( \mathcal{M}_{\text{SMoE}} \), comprises a sequence of MoE layers where each such layer contains \( M \) experts \( \{E_1, E_2, \ldots, E_M\} \). The decision to select the experts in the MoE layer on a per-token basis is made through a routing/gating function local to each MoE layer, that uses the previous layer as input. The dense model \( \mathcal{M}_{\text{Dense}} \)  which is often called the \textit{base model} or the backbone model is the equivalent MoE model when the number of experts in each MoE layer is one, which makes the routing redundant. 

Understanding how to scale sparse language models to trillions of parameters using SMoEs have been studied in models like Switch-MoE \cite{fedus2022switch} and GLAM \cite{du2022glam} where the authors scale a 5B parameter T5 model with 2048 experts and a 64B decoder model with 64 experts respectively and more recently, in similar architecture  models with less than 100B parameters \cite{xue2024openmoe,jiang2024mixtral,dai2024deepseekmoe}. In this paper, we follow the GLAM style GPT2 language model decoder architecture where we replace every alternate feedforward (FFN) layer with an MoE layer starting with the first FFN layer.
Despite the larger parameter space, \( \mathcal{M}_{\text{SMoE}} \) aims to maintain an inference computational time comparable to \( \mathcal{M}_{\text{Dense}} \). This is often quantified in terms of FLOPs (Floating Point Operations) per token in a forward pass. The goal is to achieve per token \( \text{FLOPs}(\mathcal{M}_{\text{SMoE}}) \approx \text{FLOPs}(\mathcal{M}_{\text{Dense}}) \), despite the overall larger parameter size of \( \mathcal{M}_{\text{SMoE}} \). 

Scaling laws have been studied to understand the tradeoffs between the number of experts and the base model size when studying the loss optimal configuration \cite{clark2022unified}. What has evaded such studies is the observation that the inference constraints of SMoE models are not always characterized by the model FLOPs per token \cite{huang2023towards,rajbhandari2022deepspeed}, especially in the setting where a large number of experts are used to scale the model size as studied in the Switch transformer. Unlike pretraining, LLM inference is memory-intensive and with more experts, this memory is constrained, reducing batch size and throughput, thereby increasing the cost per query. In the current regime of SMoE model architectures, the tradeoffs that come with varying the number of experts per MoE layer has not been studied in depth in the literature. 


Deploying SMoEs with hundreds of experts per layer in production with distributed GPU clusters require newer forms of parallelism. Techniques like expert parallelism \cite{aminabadi2022deepspeed, rajbhandari2022deepspeed} are necessary during inference with batched requests to overcome the per-GPU memory constraints as well as to increase throughput. In a distributed setting, this however leads to extra inter-GPU communication costs leading to slower decoding in inference and recent studies have dealt with them through MoE specific pruning techniques \cite{li2023merge,eliseev2023fast} and quantization \cite{kim2023mixture}. This begets a key question: if we keep the inference latencies in context and want to distill or prune a pretrained SMoE model to one with fewer task specific experts during inference, are there any inherent advantages of pretraining an overall larger SMoE with more experts. This motivates the following:

\begin{mdframed}[skipabove=0.2cm, skipbelow=0.2cm, innermargin=0.1cm, outermargin=0.1cm, linewidth=0.5pt, linecolor=gray]
How do we decide among the choices of having fewer or more experts in SMoE pretraining if we are later restricted to a memory budget with task specific inference?
\end{mdframed}

We study this question of the choice of experts from a task-aware post-training pruning perspective, since SMoEs inherently have one advantage over dense models when it comes pruning for inference: \textit{routing information}, that can inform decisions to remove entire expert modules in an attempt to reduce memory consumption. It exhibits forms of structured sparsity implicitly, which unlike dense model pruning, does not need any  additional sparse operator kernel specific optimization post-training for deployment when compared to pre-training. 

This question is also more fitting since as we describe later, there is a tradeoff and a Pareto frontier between three factors, when there is choice of varying the number experts (where more experts will increase overall parameter count and increase memory): (1) the model performance, (2) the latency incurred by the SMoE models during inference in distributed settings and, (3) the pretraining compute costs. Training a single SMoE model that is flexible to all these three and simultaneously optimal is difficult to achieve. While training costs are often borne by large entities,  the end consumer has to bear the inference costs, and therefore understanding these tradeoffs become necessary from a pretraining standpoint. To that end and to our knowledge, we develop the first controlled study to answer and conclude the following:

\begin{enumerate}
    \item \textbf{When is it advantageous to pretrain SMoEs with fewer experts from a task performance perspective?} Our results suggest that it is beneficial to train models with larger number of experts when we have downstream inference budgets that can handle smaller reduction ratio in expert counts, in our study, by a factor of 2 or less. On the contrary, when one is heavily memory budget constrained during inference more than at training, it is beneficial to have fewer experts in pretraining even if one decides to reduce the number of experts by a factor of 4 through task specific expert pruning later.
    \item \textbf{Can task specific expert pruning (with the intent of model compression) of larger SMoE models retain the performance benefits over similar models trained from scratch?} We find that the pruning strategy is crucial in these cases and a naive expert activation based pruning strategy hurts performance significantly. We propose a new technique {\tt UNCURL} that overcomes issues of information loss with expert pruning through expert merging and that retains the benefits of larger SMoEs.
\end{enumerate}

The focus of our work is not solely to provide new pruning algorithms.  When trying to decide the architectural elements of SMoEs in pretraining, we aim to understand what implications does ``less costly" expert pruning with the advantage of the routing information in expert layers in SMoEs have. In the quest for that, as an example, we study whether one should pretrain from scratch a dense 354 million parameter GPT2 model upscaled with 8 experts per MoE layer (1.3 billion parameters in total) over the dense model but upscaled with 64 experts per MoE layer (13B in total), when in both situations, the forward FLOPs per token remain the same, due to the nature of routing. We answer this question from the lens of token-expert activation patterns and clustering to then show how having more experts leads to better generalization for downstream tasks in certain cases but at the cost of worse results with pruning. Additionally, with large SMoEs, overfitting to downstream tasks with invdividual task finetuning has been a know problem \cite{fedus2022switch,shen2023mixture,zadouri2023pushing} and this also motivates the need to reduce the SMoE model parameters during downstream task finetuning. 


\section{Technical Preliminaries on SMoE}
\label{sec:prelim}
We start with the conventional transformer based GPT2 architecture which is the basis of all our experiments in the paper. Such a model has $N_l$ transformer layers, a hidden dimension of $d$ and context length $L$. An MoE layer takes an input representation $\mathbf{x}$ and  the sparsity arises from the setting where the input is routed to $k$ experts out of $M$ experts in a layer. Here, we consider that \( \text{MoE}(\mathbf{x}) = \sum_{i=1}^{k}\mathbf{g}_i(\mathbf{x}) \cdot \text{FFN}(\mathbf{x})  \), where $\mathbf{g}_i(\mathbf{x})$ is a learnable routing function that weights how much importance $\mathbf{x}$ gives to expert $i$. In standard SMoEs, $k$ is usually set to 1 or 2 and the manner in which the set of experts are chosen for each $\mathbf{x}$ determines whether the routing is learnt \cite{lewis2021base,shazeer2017outrageously} or whether it is static \cite{roller2021hash,gururangan2021demix}. In this paper, we restrict ourselves to the case where $k=1$, so for each MoE layer in our model, we activate one expert per input per MoE layer.  


We follow the Switch transformer \cite{fedus2022switch} style greedy routing/gating function where the router is parameterized by a variable $\mathbf{W} \in \mathbb{R}^{d \times M}$ and it produces logits $h(\mathbf{x}) = \mathbf{W}^T \mathbf{x}$ which are then normalized via a softmax distribution over the available $M$ experts at that layer. The gate-value for expert $i \in \{|M|\}$ is given by  $g_i(x) = \frac{e^{h(x)_i}}{\sum_{j=1}^{M} e^{h(x)_j}}$. In a top-$\mathbf{K}$ style routing or expert selection in an MoE layer, we sort the probabilities $g_i(x)$ and pick the top $\mathbf{K}$ values. So it goes without saying that when $\mathbf{K}=1$, the SMoE based model has virtually the same number of parameters as a dense model such that the FFN layers in the dense model are a replica of each expert in the SMoE layer. It is worth noting that even when we use one expert per MoE layer, the output of the layer is weighted by $g(\mathbf{x})$ as it ensures that the router weights are updated during backpropagation. It is worth noting that when keeping $\mathbf{K}$ fixed, increasing $M$ does not increase the per token FLOPs needed for a forward pass. Throughout the paper we use the notation {\tt 354M+Xe} as an SMoE model where the alternate FFN layers of a 354 million parameter dense GPT2 model have been upscaled to MoE layers, each having $X$ copies of the FFN module. 

An auxiliary differentiable load balancing loss is introduced for load balancing among the experts following the Switch transformer \cite{fedus2022switch}. This  balancing loss on a batch of tokens is formulated as: $\text{loss} = \alpha \sum_{i=1}^{M} f_i \cdot P_i$. Here, \( f_i \) represents the fraction of tokens dispatched to expert \( i \), defined as: $f_i = \frac{1}{|\mathbf{B}|} \sum_{x \in \mathbf{B}} \1 \{\text{argmax} \ g_i(x) = i\}$. And \( P_i \) is the fraction of the router probability allocated for expert \( i \), given by: $P_i = \frac{1}{|\mathbf{B}|} \sum_{x \in \mathbf{B}} g_i(x)$. In these equations, \( \mathbf{B} \) represents the batch of inputs, and \( \alpha \) is a scaling factor, \( \1\{\cdot\} \)  denotes the indicator function. Also, worth noting is that this nature of routing is known as ``learnt" routing \cite{dikkala2023benefits} rather than routing that depends on pre-determined domain specification \cite{sukhbaatar2024branch}.

\section{Tradeoffs in Performance, Training and Inference Costs}
\label{sec:tradeoffs}
\begin{wrapfigure}{r}{0.5\textwidth}  
\centering
\vspace{-10pt}  
\includegraphics[width=0.48\textwidth]{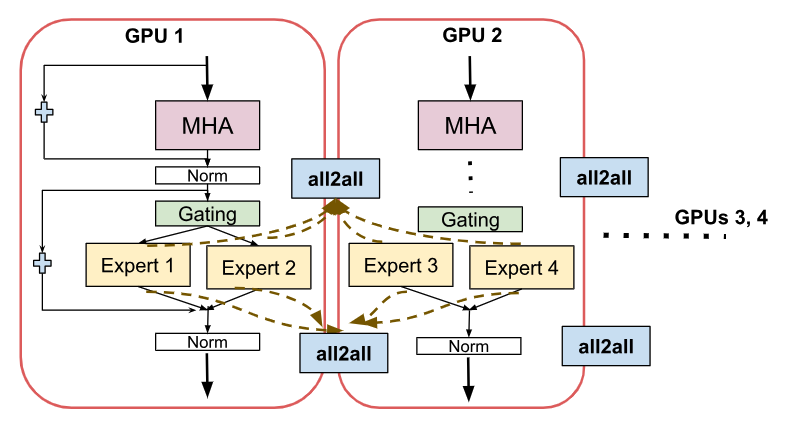}  
\vspace{-5pt}  
\caption{MoE models over a distributed setup with Expert Parallelism.}
\vspace{-10pt}  
\label{fig:smoe}
\end{wrapfigure}

As mentioned in \cite{lepikhin2020gshard}, the nature of memory increase with more experts per MoE layer necessitates the use of expert parallelism   that distributes the experts over multiple GPUs, both in training and inference. An example transformer block in a distributed setup is shown in Figure~\ref{fig:smoe}, where we consider 8 experts of an MoE layer to be distributed over 4 GPUs and the non-MoE layers are replicated across all GPUs. An expert parallel degree of 4 would mean that we place 8/4=2 experts on each GPU.
In batched training and inference, where input tokens are distributed over these GPUs, the learnt routing necessitates data movement (tokens) across GPUs since experts selected for tokens may not reside on the same device as the token. This leads to \textit{All2all} communication \footnote{https://pytorch.org/docs/stable/distributed.html} and what follows is that the latency cost is now a function of two factors: the time it takes for expert computation and the time for token to expert data movement based $All2all$ communication. 

\begin{wrapfigure}{r}{0.5\textwidth}  
\centering
\vspace{-10pt}  

\begin{minipage}{0.5\textwidth}  
    \centering
    \includegraphics[width=\textwidth]{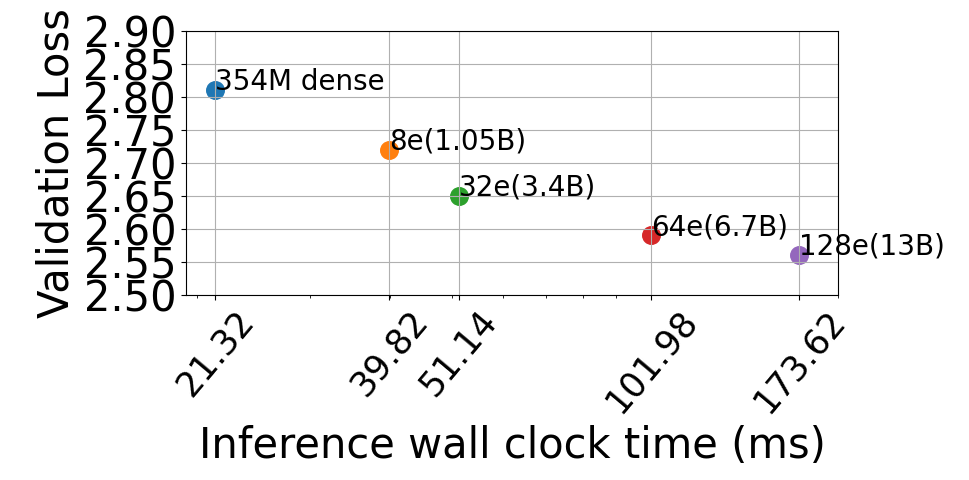}  
    \vspace{-14pt}  
    \subcaption{Loss vs Inference costs}\label{fig:val_inf_1}
\end{minipage}
\vspace{-3pt}  

\begin{minipage}{0.36\textwidth}  
    \centering
    \includegraphics[width=\textwidth]{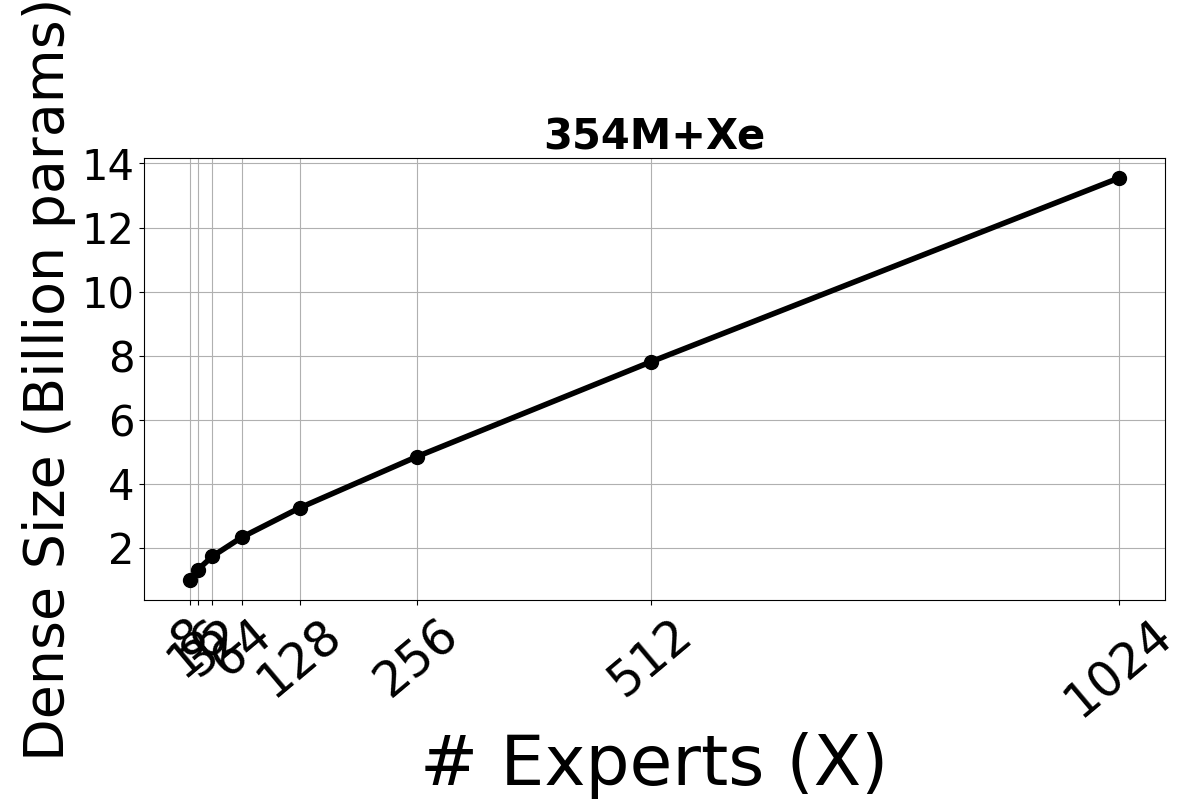}  
    \vspace{-14pt}  
    \subcaption{(Performance Equivalent) Dense Model Size when increasing \# experts}\label{fig:val_inf_2}
\end{minipage}
\caption{Tradeoffs between model performance and inference latencies}
\label{fig:val_inf_combined}
\end{wrapfigure}

To understand the tradeoffs between performance and inference costs, we start by pretraining from scratch {\tt 354M+Xe} SMoE models. We train 4 different models on the same data, setting $X$ to 8, 32, 64 and 128 experts in each MoE layer and using one expert per layer for a token.  As we increase the number of experts in each MoE layer, the total parameter count of the model increases but the number of activated parameters per token remains the same as the dense 354M in each configuration, since we use top-1 routing. The details of the datasets are described in Section \textbf{A2} of the Appendix. We use the deepspeed library\footnote{https://www.deepspeed.ai/tutorials/mixture-of-experts-nlg/} for the implementation of top-K routing in MoE layers \cite{rajbhandari2022deepspeed}, and the details of the training setup where each model is trained to 40B tokens is described in Section \textbf{A3} the Appendix. We plot this tradeoff between validation loss as the proxy for model performance vs the latency profiles in Figure~\ref{fig:val_inf_1}. Inference times for a forward pass to compute logits is profiled with eight A10 24GB GPUs placed within one node and with a batch size of 2 per GPU and 256 tokens per example, expert parallelism factor of 2 (does not matter for dense models). The inference latencies are averaged over an additional 10 forward passes for each model after discarding the first 5 runs.
Observing Figure~\ref{fig:val_inf_1}, as we increase the number of experts with the 354M backbone model, the inference latencies also increase monotonically. Observing Figure~\ref{fig:val_inf_2}, the appeal of training larger SMoEs with more experts per layer primarily stems from the empirical observation in the scaling laws proposed by \cite{clark2022unified}. As we increase {\tt X} in the {\tt 354M+Xe} SMoE models, the estimated \textit{performance equivalent} dense model size increases. But we do not proportionally increase the FLOPs per token, with {\tt X}. So, while  training a {\tt 354M+32e} (3B) model allows us to achieve better performance than a {\tt 354M+8e} (1B) model, the disadvantage lies in the relative higher inference costs of {\tt 354M+32e}. The higher inference is mostly attributed to increased $All2All$ communications with expert and data parallelism across GPUs. This makes it infeasible for consumers to pretrain multiple of these SMoE models, which they can then use flexibly depending on whether they choose inference constraints strictly or the performance. A more detailed description of the inference latency issue can be found in Section \textbf{A1} of Appendix.

We aim to determine whether entire experts in larger SMoE models can be flexibly pruned specific to each task. For instance, a {\tt 354M+8e} model is clearly faster at inference than a {\tt 354M+64e} model, even though both have the same activated parameter count. This leads us to investigate whether it is preferable to pretrain a {\tt 354M+8e} model and finetune it for each task without the need for later distillation or pruning, or if we can pretrain a {\tt 354M+64e} model and effectively prune it to a {\tt 354M+8e} model in a task-dependent manner. Moreover, we examine if the final task-pruned {\tt 354M+8e} model performs better than the one trained from scratch.

Thus, our study proposes a new algorithm focused on two aspects: (1) whether expert pruning in SMoEs can retain the performance benefits (without costly optimization based methods for structured pruning) and (2) what implications does that have when trying to understand the size of models to pretrain based on desired downstream task performances.

\section{Can expert pruning retain the performance benefits?}
\label{sec:pretraining}
\subsection{Experiment Setup}
\label{sec:pretraining_exp}
As a first test, we conduct an ablation to assess whether expert pruning can retain the benefits of the larger SMoEs. We use the same set of pretrained models described in Section~\ref{sec:tradeoffs} and for downstream task evaluations of the pretrained models, we utilize the list of tasks from SuperGLUE as used in ST-MoE \cite{zoph2022st}. However, unlike that study, we adopt a two-stage approach for task-specific evaluation as shown in the figure. In the first stage, we perform a one-shot task-dependent expert pruning and then finetune the pruned model on instruction-output pairs (separate from the task in context). This prefinetuning stage helps in evaluation since our models have not been pretrained on enough tokens to adapt easily to downstream instruction-based tasks. In the second stage, we further finetune the pruned and prefinetuned model on the training splits of the tasks individually. For non-pruned SMoE models, the first stage is similar, but without the pruning step.
\begin{wrapfigure}{r}{0.4\textwidth}  
\centering
\vspace{-10pt}  
\includegraphics[width=0.42\textwidth]{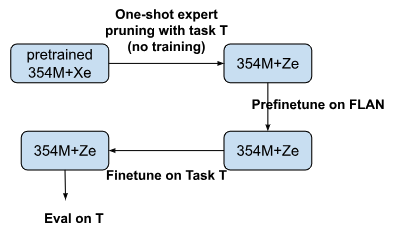}  
\vspace{-10pt}  
\caption{Pruning and Evaluation}
\label{fig:workflow}
\end{wrapfigure}
For the prefinetuning in the first stage, we use the FLAN dataset, introduced in \cite{wei2021finetuned} for the supervised instruction-output pairs. We exclude the tasks from SuperGLUE that we use for evaluation that were also present in FLAN namely, COPA, BOOLQ, RTE, CommitmentBank, RECORD and WSC from the original FLAN mixture since we define our own train and validation splits for these tasks in the second stage for task specific finetuning. Additionally, we add a portion of the pretraining dataset to the FLAN dataset for the first stage to mitigate catastrophic forgetting known to affect multi-stage finetuning.

\begin{table*}[!tbp]
\centering
\small
\resizebox{\textwidth}{!}{
\begin{tabular}{p{2.8cm}|p{1.3cm}|p{1cm}|p{1cm}|p{1cm}|p{1cm}|p{1.3cm}|p{1cm}}
\toprule
\textbf{Model} & \textbf{Total Params} & \textbf{RTE} & \textbf{BoolQ} & \textbf{COPA} & \textbf{CB} & \textbf{ReCORD} & \textbf{WSC}    \\ \midrule
\textbf{354M} & 354M & 64.21 & 65.33 & 56.61 & 51.32 & 63.14 & 65.35  \\ \midrule
\textbf{354M+8e} & 1.05B & 66.03  & 67.35 & 58.4 & 50.87 & 65.17 & 68.16 \\ \midrule
\textbf{354M+32e} & 3.4B & 66.86  & 68.27 & 59.13  & 51.86 & 67.75 & 70.87  \\
\textbf{354M+(32e $\to$8e)}$^\dagger$ & 1.05B  & 64.86  & 66.71 & 58.12 & 49.53 & 64.77 & 67.59 \\ \midrule
\textbf{354M+64e} & 6.7B & 67.64  & 69.09 & 60.05  & \textbf{52.17} & 68.85 & 71.68  \\ 
\textbf{354M+(64e $\to$32e)}$^\dagger$ & 3.4B & 66.36  & 68.21  & 58.74  & 50.09  & 66.86 & 69.43  \\ \midrule
\textbf{354M+128e} & 13B & \textbf{68.58}  & \textbf{70.13} & \textbf{61.3}  & 52.04 & \textbf{69.83} & \textbf{72.18}  \\
\textbf{354M+(128e $\to$64e)}$^\dagger$ & 6.7B & 66.91  & 68.22  & 58.31  & 50.98 & 67.22 & 70.09\\ 
\bottomrule
\end{tabular}
}
\caption{$^\dagger$The notation {\tt 354M+(32e->8e)} means that the pretrained {\tt 354M+32e} model was pruned to a {\tt 354M+8e} SMoE in the manner shown in Figure~\ref{fig:workflow}. Bold values denote the best performance across all.}
\label{tab:freq_prune}
\end{table*}
\subsection{Observations}
For the one-shot pruning strategy, we use an expert activation frequency method to prune an {\tt354M+Xe} model to an {\tt354M+Ye} model, where $X > Y$. For each layer, we calculate the frequency of each expert's selection for the tokens in the training split of the task. We then retain the top $Y$ experts in each layer based on these frequencies. Table~\ref{tab:freq_prune} shows the results of the pruned models compared to their equivalent SMoEs trained from scratch. The direct observation is that all pruned models—{\tt 354M+(32e$\to$8e)}, {\tt 354M+(64e$\to$32e)}, and {\tt 354M+(128e$\to$64e)}—fail to recover the performance gains after pruning when compared to the equivalent SMoE models trained from scratch. This occurs despite additional finetuning after pruning, indicating that the loss of token-expert routing information in the pruned experts makes effective model compression through expert pruning challenging. This phenomenon is consistent across all the SuperGLUE tasks we used, demonstrating its generality regardless of the specific task (noting that the pruning in the first stage is task-data dependent).

\section{Pruning with Clustering and Merging}
\label{sec:expert_merging}

We hypothesize that the one-shot pruning strategy discussed previously leads to an irrecoverable loss of capacity and knowledge, even with additional two-stage finetuning. With recent advances in model merging techniques \cite{wortsman2022model,ainsworth2022git,yadav2024ties}, we propose that merging in the expert weight subspace may mitigate the degradations caused by pruning. Note that this method is a substitute to the one-shot pruning technique mentioned above. ModuleFormer \cite{shen2023moduleformer} introduces a mutual information-based loss during pretraining to uniformly distribute tokens across experts and a load concentration loss during task finetuning to achieve the opposite effect, allowing for expert pruning post-finetuning. Rather than relying on regularization for implicit load concentration of experts, we use clustering per MoE layer for the one-shot pruning stage.  While there have been attempts at expert merging as an extension of model merging techniques during training \cite{muqeeth2023soft}, we instead use an offline non-parametric clustering technique as \textbf{k}-means to obtain the reduction in the expert space. The expert clustering is conceptualized as an offline step, performed on the pretrained checkpoints. We do not use any additional supervised training in this stage, and instead use the finetuning data mixture for the clustering. We then do a weighted average of the expert parameters in each cluster to form one expert per cluster, a techniques we term \textbf{\textit{cluster-merging}}. Post the cluster-merging stage that produces a reduced set of experts, we finetune the models with the task data as shown in Figure~\ref{fig:workflow}.

In what follows, we apply the clustering algorithm for each layer separately. The study in \cite{li2023merge} demonstrated a way to merge redundant experts by comparing the cosine distances between router logits. One of the nuances of the expert grouping technique in that study is that, since the method relies on identifying groups based on all global experts in the model, it does not guarantee a reduction to $D$ experts per layer starting from $Z$, but only ensures an average $D$ groups per layer. This makes it inconvenient to compare the resulting pruned SMoE to an equivalent model trained from scratch. Note that unlike distillation as done in \cite{fedus2022switch}, which does not explicitly control how the knowledge of experts are grouped, our method controls the expert grouping based on information flow and has the flexibility to be adaptive to different task data during finetuning which is crucial since the expert activations depend on the task data. Next, we describe the algorithm for this pruning process.


\subsection{\texorpdfstring{UnCuRL: \underline{Un}ified \underline{C}l\underline{u}stering Based on \underline{R}outer \underline{L}ogits (Laywerise)}{UnCuRL: Unified Clustering Based on Router Logits (Laywerise)}}

\begin{wrapfigure}{R}{0.5\textwidth}
\vspace{-5ex}
\resizebox{0.5\textwidth}{!}{
\begin{minipage}[!h]{0.6\textwidth}
       \begin{algorithm}[H]
        \caption{UNCURL for MoE}
        \textbf{Input:} $Z$, Multi-head attention output $X \in \mathbb{R}^{b \times t \times d_{\text{model}}}$, Router weights $W_l \in \mathbb{R}^{d_{\text{model}} \times Z}$, target $D$ experts, $\mathcal{T}$ \\
        \textbf{Output:} Reduced set of experts $\{E_{1}, \ldots, E_{D}\}$
        \begin{algorithmic}[1]
        \State Compute router logits $\mathbf{H}_l = W_l @ X \in \mathbb{R}^{b \times t \times Z}$ for a batch of $b$ samples, $t$ tokens per sample with dataset $\mathcal{T}$
        \For{each expert pair $(i, j)$ where $1 \leq i, j \leq Z$}
            \State Compute $\mathbf{S}[i, j] = \mathbf{S}[j, i] = \frac{\mathbf{H}[:, :, i]^\top \mathbf{H}[:, :, j]}{\|\mathbf{H}[:, :, i]\| \|\mathbf{H}[:, :, j]\|}$
            \State S := (1.0 + S) / 2.0
        \EndFor
        \State Apply spectral clustering to obtain eigenvectors $\mathbf{F} \in \mathbb{R}^{Z \times D}$ of the graph laplacian of $\mathbf{S}$
        \State $\mathbf{C} \leftarrow \text{\tt {K-means}} (\mathbf{F}, \text{num\_clusters=D})$ \text{// cluster labels} $\in \mathbb{R}^{Z \times 1}$

        \For{each cluster $d$ in $\{1, \ldots, D\}$}
        \State $\mathcal{G}_d$ $\leftarrow$ $\{E_y | C[y] == d\}$ \text{// expert set}
        \State $\mathcal{G}'_d$ $\leftarrow$ $[\ ]$
        \State Identify expert $J \in \mathcal{G}_d$ as the most frequently selected expert on $\mathcal{T}$, add $E_J$ to $\mathcal{G}'_d$ 
        \For{each expert $y \in \mathcal{G}_d$, $y \neq J$}
           \State $E'_y$ $\leftarrow$ \text{\tt {Permutation\_Align}}($E_j$, $E_y$)
           \State Add $E'_y$ to $\mathcal{G}'_d$
        \EndFor
        \State $E'_d \leftarrow \text{\tt {WeightedAverage}}(\mathcal{G}'_d)$
        \EndFor
        
        \Return $\{E'_{1}, \ldots, E'_{D}\}$
        \end{algorithmic}
        \label{alg:uncurl}
        \end{algorithm}
     \end{minipage}
}
\vspace{-3ex}
\end{wrapfigure}

We consider an SMoE model comprising \( Z \) experts. Let \( \mathcal{T} \) represent the  task data that we use for determining the logits from the router of layer $i$. We perform the following process individually for each layer of the SMoE model and at the risk of generalizing, we remove any layer specific identifiers in the notations where possible. The router in layer $i$ parameterized by $\textbf{W}_i$ generates a logits vector \( \mathbf{g} \) of dimension \( Z \), where each element \( g_{z} \) corresponds to the router's output for expert \( m \in \{|M|\} \). To get an SMoE model with reduced expert count of $D$ experts per layer, the goal is to cluster the \( Z \) experts into $D$ clusters based on expert similarity, as inferred from the logit responses across the \( |\mathcal{T}| \) data. The algorithm for {\tt UNCURL} for each MoE layer $l$ is outlined in Algorithm~\ref{alg:uncurl}. We now describe the algorithm briefly.

For a layer with $Z$ experts, we construct an expert similarity matrix \( \mathbf{S} \) $\in \mathbb{R}^{Z \times Z}$ (Lines 2-4) where each element \( S_{ij} \) measures the similarity between experts \( i \) and \( j \), calculated based on their logits across all data points. We apply spectral clustering on \( \mathbf{S} \)  that transforms it into a lower-dimensional space using the eigenvalues and eigenvectors of the Laplacian matrix \( \mathbf{L} \) \cite{ng2001spectral}. We then apply $K$-means clustering to these eigenvectors to partition the experts into \( D \) clusters. Inspired by the model merging process used in \cite{wortsman2022model}, we perform weighted averaging of the experts ({\tt {WeightedAverage()}}) in each cluster to form one expert per cluster post aligning them per cluster (Lines 12-16). The weights in the averaging are the activation frequencies of the experts on task $T$. When finetuning this cluster-merged model further, we randomly reinitialize the weights of the router accordingly with appropriate dimensions reflecting $D$ experts per layer. It is important to note that more sophisticated weight merging strategies cited in \cite{yadav2024ties, ainsworth2022git} are not in competition with our method but are complementary; our technique of activation frequency weighted merging could be substituted with these approaches. 

Prior to cluster-merging the experts within a layer, we align expert weight permutations to prevent suboptimal fusion of mismatched neurons following the work done in \cite{ainsworth2022git} and adopted in \cite{li2023merge}. In our case, since we consider experts in the same layer, the input and output spaces are similar. Let \( W_{\text{in}} \) and \( W_{\text{out}} \) represent input and output layer weight matrices (two layer feedforward network of the experts), and \( \mathbf{x} \) be the input vector. Denoting the activation function as \( \text{act}(\cdot) \), the network mapping \( F: \mathbf{x} \to W_{\text{out}}(\text{act}(W_{\text{in}}\mathbf{x})) \) expresses expert operations. To align experts without altering their functionality, we utilize the weight matching optimization technique  from \cite{li2023merge}. An optimal permutation matrix \( \mathbf{P} \) is identified identified for experts \( E_i \) and \( E_j \) with weight matrices \( \mathbf{W}_i \) and \( \mathbf{W}_j \).  We note that, since we align all the experts in one MoE layer  to the most activated expert $E_j$ in that layer (Line 11), we do not need to align the rest of the weights of the model, as the weights of $E_j$ is unchanged and that is our reference weight for that layer. The optimization minimizes \( \ell_2 \) distance between corresponding permuted weights \( \mathbf{W}'_i \) and \( \mathbf{W}_j \), facilitating effective merging. This leads to the following optimization which constitutes a linear assignment problem: 
$\mathrm {argmax}_{\mathtt P} \left\langle \mathtt W_{\text {in}}^{(E_i)}, \mathtt P\mathtt W_{\text {in}}^{(E_j)} \right\rangle_{\text F} + \left\langle \mathtt W_{\text {out}}^{(E_i)}, \mathtt W_{\text {out}}^{(E_j)}\mathtt P^{\text T}\right\rangle_{\text F}$ which can be solved using the Hungarian Algorithm \footnote{https://docs.scipy.org/doc/scipy/reference/generated/scipy.optimize.linear\_sum\_assignment.html}. This procedure is captured by \text{\tt {Permutation\_Align}}() in Line 13 of the algorithm.

\subsection{Time Complexity}
The time complexity of the {\tt UnCuRL} algorithm is primarily dominated by the similarity computation, which has a time complexity of \(\mathcal{O}(Z^2 \cdot |\mathcal{T})|\). Spectral clustering and the linear assignment problem both have a time complexity of \(\mathcal{O}(Z^3)\). K-means clustering contributes \(\mathcal{O}(I \cdot Z \cdot D)\), where \(I\) is the number of iterations, \(D\) is the number of clusters. For large-scale data where \(\mathcal{|T|} \gg Z\), the similarity computation dominates the overall time complexity.

\section{Results}
\label{sec:finetuning}
In this section, we evaluate the results of the cluster-merging technique introduced in Section~\ref{sec:expert_merging}. Following the workflow depicted in Figure~\ref{fig:workflow} for task-specific pruning and evaluation, we replace the greedy pruning in Stage 1, as described in Section~\ref{sec:pretraining_exp}, with the new cluster-merging method.

\subsection{Performance of SMoE models on downstream tasks}
We first discuss the results of applying the expert clustering and merging on downstream tasks in this section. Recall that our goal is to understand whether experts in larger SMoE models can be flexibly pruned to a reduced set while retaining benefits over equivalent smaller SMoEs trained from scratch. We first compare the results of the SMoE models as we scale them with more experts per MoE layer in the 354M model. From the results in Table~\ref{tab:uncurl}, we see that for all tasks except Commitment Bank (CB), the performance of the SMoE models improves on the tasks as we scale the models with more experts. This is probably due to the fact that CB has roughly 300 examples for training and even with multi-task finetuning, the larger {\tt 354M+128e} model overfits. When comparing the largest SMoE model we trained  {\tt 354M+128e} with the base dense model of 354M, we observe that we get performance improvements of +7.3\% on the BoolQ task and 6.8\% on RTE which are among the highest improvements we observe throughout.


\begin{table*}[!tbp]
\centering
\small
\resizebox{\textwidth}{!}{
\begin{tabular}{p{2.8cm}|p{1.3cm}|p{1.2cm}|p{1.3cm}|p{1.2cm}|p{1.2cm}|p{1.3cm}|p{1.2cm}}
\toprule
\textbf{Model} & \textbf{Total Params} & \textbf{RTE} & \textbf{BoolQ} & \textbf{COPA} & \textbf{CB} & \textbf{ReCORD} & \textbf{WSC}    \\ \midrule
\textbf{354M+8e} & 1.05B & 66.03  & 67.35 & 58.4 & 50.87 & 65.17 & 68.16 \\ \midrule
\textbf{354M+32e} & 3.4B & 66.86  & 68.27 & 59.13  & 51.86 & 67.75 & 70.87  \\
\textbf{354M+(32e$\to$8e)}$^\dagger$ & 1.05B  & 66.48 \color{green}{(+)}  & 69.13 \color{green}{(+)} & 58.96  \color{green}{(+)} & 50.65 \color{red}{(-)} & 66.34 \color{green}{(+)} & 69.86 \color{green}{(+)} \\ \midrule
\textbf{354M+64e} & 6.7B & 67.64 & 69.09  & 60.05 & \textbf{52.17} & 68.85 & 71.68  \\
\textbf{354M+(64e$\to$8e)}$^\dagger$ & 1.05B & 66.17 \color{green}{(+)}  & 67.27 \color{red}{(-)} & 58.11 \color{red}{(-)} & 50.16 \color{red}{(-)} & 65.01 \color{red}{(-)}& 68.11 \color{red}{(-)} \\ 
\textbf{354M+(64e$\to$32e)}$^\dagger$ & 3.4B & 67.14 \color{green}{(+)}  & 68.86 \color{green}{(+)} & 59.96 \color{green}{(+)} & 51.94 \color{green}{(+)} & 67.96 \color{green}{(+)} &  69.94 \color{red}{(-)}  \\ \midrule
\textbf{354M+128e} & 13B & \textbf{68.58} & \textbf{70.13} & \textbf{61.3} & 52.04 & \textbf{69.83} & \textbf{72.18} \\
\textbf{354M+(128e$\to$8e)}$^\dagger$ & 1.05B & 62.91  \color{red}{(-)}& 66.93 \color{red}{(-)} & 56.56  \color{red}{(-)}& 49.31  \color{red}{(-)}& 64.32  \color{red}{(-)} & 64.98  \color{red}{(-)}\\ 
\textbf{354M+(128e$\to$32e)}$^\dagger$ & 3.4B & 65.17 \color{red}{(-)} & 67.17 \color{red}{(-)}  & 59.04  \color{red}{(-)} & 50.17  \color{red}{(-)} & 67.02  \color{red}{(-)} & 71.04 \color{green}{(+)}  \\
\textbf{354M+(128e$\to$64e)}$^\dagger$ & 6.7B & 67.73 \color{green}{(+)}  & 69.28 \color{green}{(+)} & 60.82 \color{green}{(+)} & 51.81 \color{red}{(-)}  & 68.91 \color{green}{(+)} & 71.73 \color{green}{(+)} \\ \bottomrule
\end{tabular}
}
\caption{Accuracy on the validation sets of the SuperGLUE tasks after applying the {\tt UNCURL} algorithm. $^\dagger$The notation {\tt 354M+(32e->8e)} means that the {\tt UNCURL} algorithm was applied on the pretrained {\tt 354M+32e} model and cluster-merged to 8 experts per MoE layer prior to the finetuning stage. Bold values denote the best performance across all models. \textcolor{green}{(+)} denotes that the reduced model performs better than the equivalent smaller SMoE with equal experts and  \textcolor{red}{(-)} denotes worse performance. For example, for the  {\tt 354M+(32e->8e)} on RTE task, 66.48 \textcolor{green}{(+)} means the results for this model is better than {\tt 354M+8e} for that task.}
\label{tab:uncurl}
\end{table*}

\subsection{Impact of Cluster-Merging experts}
In the following discussion, we explore the principal results of the {\tt UNCURL} algorithm detailed in Section~\ref{sec:expert_merging}. Initially, we compare the \texttt{354M+(32e->8e)} model with the \texttt{354M+8e} model, both maintaining identical parameter counts. The findings indicate that except for the CB task, the reduced \texttt{354M+(32e->8e)} model consistently outperforms the \texttt{354M+8e} model, thus suggesting that performance enhancements can be achieved despite a reduction in the number of experts, compared to smaller SMoE models. Moreover, in the RTE task, the \texttt{354M+(64e->32e)} model slightly surpasses the \texttt{354M+32e} model. Interestingly, the \texttt{354M+(64e->8e)} model shows underperformance on most tasks relative to the \texttt{354M+8e} model.

When we compare the merged models of { \tt 354M+128e } variants, we find that both the { \tt 354M+(128e->8e)  } and { \tt 354M+(128e->32e)  } perform worse than the { \tt 354M+32e  } models on all tasks. This suggests that there is an impact of \textit{over}-reduction of the experts and that the pruning ratio plays an important role. A further understanding of a hybrid way to select top experts greedily by expert activation across some layers along with our cluster-merging technique for other layers would be an interesting direction to explore.

When comparing the results with the variants subjected to the greedy one-shot pruning strategy in Table~\ref{tab:freq_prune}, we observe that our proposed method can effectively prune the SMoEs by a factor of 2 for most tasks in the \texttt{354M+64e} and \texttt{354M+128e} variants, and by a factor of 4 for the \texttt{354M+32e} variant. This indicates that advanced cluster-merging techniques are essential for task-specific pruning and it allows flexibility of larger models to be later pruned to smaller models, yet be able to retain the performance benefits for tasks.

\subsection{Cluster Visualization}

We next analyze the results of applying the offline expert clustering algorithm {\tt UNCURL} and we show the clustering outputs in the below figure. We apply the t-sne visualization of the eigenvector matrix \( 
\mathbf{F} \) for each MoE layer in the {\tt 354M+64e} model. Note that the results of the clustering is obtained prior to any further finetuning of the resulting reduced model after expert cluster-merging. 
\begin{wrapfigure}{r}{6.5cm}  
\centering
\vspace{-10pt}  
\begin{minipage}{7.5cm}  
\includegraphics[width=6.5cm, height=5cm]{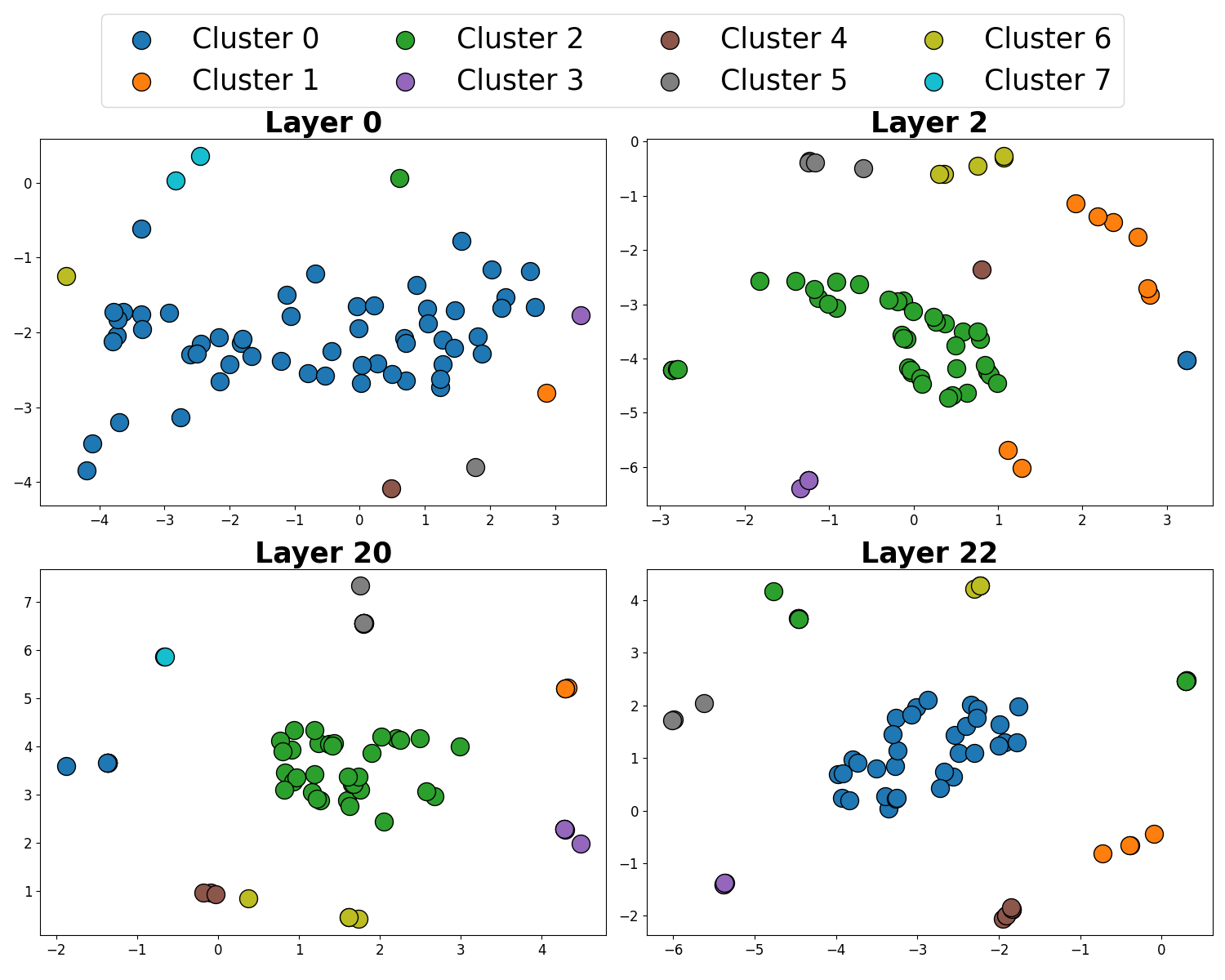}  
\label{fig:clusters}
\end{minipage}
\end{wrapfigure}
The top two images in the plot shows the clustering results from the first two MoE layers in the model and the bottom two images show the clustering results on the last two MoE layers. 
We observe that there are no visible distinct clusters for the first MoE layer which also suggests that we can skip pruning the first MoE layer for further gains. But what is noticeable is the presence of one or few distinct clusters in the rest of the layers which also emphasizes on the nature of expert specialization and possible redundancies in the expert space. Based on this, we hypothesize that specifically, for the first MoE layer, selecting the top $D$ experts greedily based on activation frequency instead of cluster-merging will yield better results and we leave this as future work.


\subsection{Comparing Baselines} \label{sec:baselines}

\subsubsection{Frequency Based Merging}
As a simple baseline, we reduce $N$ experts to a target count $K$ per layer in the following manner, where this is applied per MoE layer: we select the top $K$ experts based on frequency of usage.
\begin{wraptable}{r}{0.5\textwidth} 
\centering
\vspace{-10pt}  
\small
\begin{tabular}{@{}lccc@{}}
\toprule
\textbf{Model} & \textbf{RTE} & \textbf{BoolQ} & \textbf{COPA} \\ \midrule
354M+8e & 66.03 & 67.35 & 58.4 \\ \midrule
354M+(32e$\to$8e) $\dagger$ & 66.48 & 69.13 & 58.96 \\
\parbox[t]{3.3cm}{354M+(32e$\to$8e)*} 
& \color{red} 64.54 & \color{red}65.12 & \color{red}56.19   \\ \bottomrule
\end{tabular}
\caption{* denotes reduction following the frequency based merging and $\dagger$ denotes {\tt UNCURL}.}
\label{tab:local}
\end{wraptable}
We then label the rest of the $N-K$ experts in the same layer to either of these $K$ experts based on the similarity metric described in Lines 2-4 of Algorithm~\ref{alg:uncurl}. Note that there is no clustering per se in this situation. After the labeling, we follow the same procedure as Algorithm~\ref{alg:uncurl} to align the $N-K$ experts to their respective matched expert and then perform a weighted average of each group. From the results in Table~\ref{tab:local}, we can observe, the performance degrades, showing the nuances of using frequency based pruning in the top-1 case where merging based on grouping is not superior to our method.

\subsubsection{MC-SMOE}
\begin{wraptable}{r}{0.5\textwidth} 
\centering
\vspace{-10pt}  
\small
\begin{tabular}{@{}lccc@{}}
\toprule
\textbf{Model} & \textbf{RTE} & \textbf{BoolQ} & \textbf{COPA} \\ \midrule
354M+8e & 66.03 & 67.35 & 58.4 \\ \midrule
354M+(32e$\to$8e) $\dagger$ & 66.48 & 69.13 & 58.96 \\
\parbox[t]{3.3cm}{354M+(32e$\to$8e)*} 
& \color{green} 66.91 & \color{red}67.83 & \color{green}59.01   \\ \midrule
\parbox[t]{3.3cm}{354M+(64e$\to$32e)$\dagger$} 
&  67.14 & 68.86 & 59.96   \\
\parbox[t]{3.3cm}{354M+(64e$\to$32e)*}  
& \color{red} 67.07 & \color{red}67.96 & \color{red}58.62 \\ \midrule
\parbox[t]{3.3cm}{354M+(128e$\to$64e)$\dagger$} & 67.73 & 69.28 & 60.82   \\
\parbox[t]{3.3cm}{354M+(128e$\to$64e)*} & \color{red} 67.41 & \color{red}68.47 & \color{red}59.14   \\ \bottomrule
\end{tabular}
\caption{* denotes merging following the method of MCSMOE \cite{li2023merge}  whereas $\dagger$ denotes {\tt UNCURL}. Green denotes better performance compared to {\tt UNCURL} and red otherwise. }
\label{tab:merg_comp}
\end{wraptable}
One goal derived from our problem statement is to avoid costly retraining methods like distilling larger SMoE models into dense or smaller models from scratch, and favoring pruning instead to measure the comparisons between pruned larger SMoEs and equivalent smaller SMoEs. This preference is due to the goal of minimizing the significant computational demands of training models from the ground up as would be more common in practical settings. The study most related to our work is the merging technique MC-SMOE presented in \cite{li2023merge}. There are two main differences between their algorithm for pruning (Algorithm 1 in \cite{li2023merge} and {\tt UNCURL}: (1) their approach first identifies globally dominant experts across layers based on frequency of activation, to which other experts are then merged into. Instead, we focus on local expert clusters within each layer without any notion of activation to identify clusters instead focusing on router logits, and (2) their algorithm does not ``cluster" experts but uses similarity based grouping to reduce the expert space, which contrasts our method.

We have adapted their algorithm to better suit our pipeline. Specifically, we incorporate their global expert identification and permutation alignment strategies without adopting their additional post-merging compression or the use of knowledge distillation auxiliary loss. Note that as mentioned earlier, the method of \cite{li2023merge} does not guarantee the same number of experts per layer.  In their case,  { \tt 354M+(128e->64e)  } means that we set the number of dominant experts to 64 and their method ensures an average of 64 experts per layer. When we compare their results in Table~\ref{tab:merg_comp}, we find that for the { \tt 354M+(32e->8e)  } case, their method performs comparably to ours, even surpassing us on RTE. But when the models have more experts, their methods perform substantially worse than our proposed approach. One of the reasons we hypothesize is that globally identifying dominant experts based on activation frequency may be suboptimal in the presence of larger number of experts.

\section{Related Work}

SMoE models have witnessed research highlights from many directions including better routing techniques \cite{chi2022representation,zhou2022mixture,dai2022stablemoe,fedus2022review}, better infrastructure development for reducing communications with expert parallelism \cite{gale2023megablocks}, scaling laws for sparse models \cite{clark2022unified} and the nature of expert specialization \cite{gururangan2021demix, li2022branch}. Our study relates to task-specific distillation as discussed in \cite{kudugunta2021beyond} and SMoE model reduction through expert merging as explored in \cite{li2023merge}, but our controlled study focuses on comparisons that inform better pretraining decisions, instead of mere compression techniques. \cite{clark2022unified} studies the choice of \# experts from a loss optimal scaling perspective, whereas we study them from an inference perspective. In our paper, we focus on sparsity that comes along with gating which is contrary to other forms of sparse model training \cite{jaszczur2021sparse}.  Instead of starting with a large model and pruning, sparsity is enforced from the beginning through the gating. When it comes to understanding how experts could be further utilized for multi-task finetuning, there have been studies \cite{kim2021scalable} which have shown the importance of multi-task MoEs with multiple experts. The recent studies on instruction finetuning \cite{zadouri2023pushing} have also provided a way to mitigate the issues of SMoEs showing overfitting patterns for some downstream tasks \cite{fedus2022switch}.
In the area of language modeling, most  studies do not attempt at forcing expert specialization explicitly in the pretraining stage. The Demix studies \cite{gururangan2021demix} groups inputs by discrete domains during pretraining, but with real world internet scale web data, it is almost impossible to map inputs to the nature of continuously changing domains during pretraining. Instead, the nature of learning to route allows such SMoE models to implicitly learn the structure of the data. The recent study (Theroem 2, \cite{dikkala2023benefits}) shows that when you consider a setting with mixture of $k$ Gaussians $g_c$ and the mixture distribution defined by $\sum_{c=1}^K w_cg_c$, $K$ denoting the clusters, $w_c=\frac{1}{K}$ denoting the mixture weights, the router will learn to route examples according to the cluster they belong to.

\section{Conclusion and Future Work}

We attempt at understanding whether the inference overheads that SMoEs with larger number of experts come along with, can be mitigated with pruning techniques that does not require retraining from scratch. Alongside, we discuss what implications such comparisons, not existing in the literature before, can have implications on deciding what models to pretrain. We observe how having more experts can be an advantage in terms of pretraining performance but more costly when it comes to inference. We then attempt to answer the question of targeted expert count reduction as we develop an algorithm that would allow us to prune the SMoE models while still retaining performance improvements. Our algorithm is flexible and could be adapted to any set of task mixtures and opens up new avenues of understanding how large SMoEs can be adapted towards downstream task specific reduction without additional latency overheads.

In this work, we did not study the nature of expert specializations to determine redundancy in the expert space which is major limitation of our work. A future direction then is to use parametric clustering techniques that decides the clusters based on domain specialization like code \cite{ni2023lever}, tables \cite{sarkar2023testing}, math \cite{ahn2024large} and multilingual aspects \cite{sarkar2022parameter} among others.  A second future work is to be able to understand quantization as a means to further competitively reduce the size of the models beneficial to inference and compare that to our method.  Regarding pruning, we  need to theoretically understand what is the maximum compression rate to retain the better performance compared to an SMoE model with X experts trained from scratch. Our results show that at larger number of experts (128), getting to more than a pruning factor of 2 results in worse performance whereas for models with fewer experts (32), we are able to prune to 8 per layer without degradation, suggesting higher pruning ratio. This suggests that the ratio of non-MoE to MoE layer parameters, as well as the size of each expert would play as factors in determining the pruning factor and this opens up further studies along this direction.

\bibliographystyle{unsrtnat}
\bibliography{main}

\begin{thebibliography}{46}
\providecommand{\natexlab}[1]{#1}
\providecommand{\url}[1]{\texttt{#1}}
\expandafter\ifx\csname urlstyle\endcsname\relax
  \providecommand{\doi}[1]{doi: #1}\else
  \providecommand{\doi}{doi: \begingroup \urlstyle{rm}\Url}\fi

\bibitem[Jacobs et~al.(1991)Jacobs, Jordan, Nowlan, and Hinton]{jacobs1991adaptive}
Robert~A Jacobs, Michael~I Jordan, Steven~J Nowlan, and Geoffrey~E Hinton.
\newblock Adaptive mixtures of local experts.
\newblock \emph{Neural computation}, 3\penalty0 (1):\penalty0 79--87, 1991.

\bibitem[Shazeer et~al.(2017)Shazeer, Mirhoseini, Maziarz, Davis, Le, Hinton, and Dean]{shazeer2017outrageously}
Noam Shazeer, Azalia Mirhoseini, Krzysztof Maziarz, Andy Davis, Quoc Le, Geoffrey Hinton, and Jeff Dean.
\newblock Outrageously large neural networks: The sparsely-gated mixture-of-experts layer.
\newblock \emph{arXiv preprint arXiv:1701.06538}, 2017.

\bibitem[Fedus et~al.(2022{\natexlab{a}})Fedus, Zoph, and Shazeer]{fedus2022switch}
William Fedus, Barret Zoph, and Noam Shazeer.
\newblock Switch transformers: Scaling to trillion parameter models with simple and efficient sparsity.
\newblock \emph{The Journal of Machine Learning Research}, 23\penalty0 (1):\penalty0 5232--5270, 2022{\natexlab{a}}.

\bibitem[Du et~al.(2022)Du, Huang, Dai, Tong, Lepikhin, Xu, Krikun, Zhou, Yu, Firat, et~al.]{du2022glam}
Nan Du, Yanping Huang, Andrew~M Dai, Simon Tong, Dmitry Lepikhin, Yuanzhong Xu, Maxim Krikun, Yanqi Zhou, Adams~Wei Yu, Orhan Firat, et~al.
\newblock Glam: Efficient scaling of language models with mixture-of-experts.
\newblock In \emph{International Conference on Machine Learning}, pages 5547--5569. PMLR, 2022.

\bibitem[Xue et~al.(2024)Xue, Zheng, Fu, Ni, Zheng, Zhou, and You]{xue2024openmoe}
Fuzhao Xue, Zian Zheng, Yao Fu, Jinjie Ni, Zangwei Zheng, Wangchunshu Zhou, and Yang You.
\newblock Openmoe: An early effort on open mixture-of-experts language models.
\newblock \emph{arXiv preprint arXiv:2402.01739}, 2024.

\bibitem[Jiang et~al.(2024)Jiang, Sablayrolles, Roux, Mensch, Savary, Bamford, Chaplot, Casas, Hanna, Bressand, et~al.]{jiang2024mixtral}
Albert~Q Jiang, Alexandre Sablayrolles, Antoine Roux, Arthur Mensch, Blanche Savary, Chris Bamford, Devendra~Singh Chaplot, Diego de~las Casas, Emma~Bou Hanna, Florian Bressand, et~al.
\newblock Mixtral of experts.
\newblock \emph{arXiv preprint arXiv:2401.04088}, 2024.

\bibitem[Dai et~al.(2024)Dai, Deng, Zhao, Xu, Gao, Chen, Li, Zeng, Yu, Wu, et~al.]{dai2024deepseekmoe}
Damai Dai, Chengqi Deng, Chenggang Zhao, RX~Xu, Huazuo Gao, Deli Chen, Jiashi Li, Wangding Zeng, Xingkai Yu, Y~Wu, et~al.
\newblock Deepseekmoe: Towards ultimate expert specialization in mixture-of-experts language models.
\newblock \emph{arXiv preprint arXiv:2401.06066}, 2024.

\bibitem[Clark et~al.(2022)Clark, De~Las~Casas, Guy, Mensch, Paganini, Hoffmann, Damoc, Hechtman, Cai, Borgeaud, et~al.]{clark2022unified}
Aidan Clark, Diego De~Las~Casas, Aurelia Guy, Arthur Mensch, Michela Paganini, Jordan Hoffmann, Bogdan Damoc, Blake Hechtman, Trevor Cai, Sebastian Borgeaud, et~al.
\newblock Unified scaling laws for routed language models.
\newblock In \emph{International Conference on Machine Learning}, pages 4057--4086. PMLR, 2022.

\bibitem[Huang et~al.(2023)Huang, Ardalani, Sun, Ke, Lee, Sridhar, Bhosale, Wu, and Lee]{huang2023towards}
Haiyang Huang, Newsha Ardalani, Anna Sun, Liu Ke, Hsien-Hsin~S Lee, Anjali Sridhar, Shruti Bhosale, Carole-Jean Wu, and Benjamin Lee.
\newblock Towards moe deployment: Mitigating inefficiencies in mixture-of-expert (moe) inference.
\newblock \emph{arXiv preprint arXiv:2303.06182}, 2023.

\bibitem[Rajbhandari et~al.(2022)Rajbhandari, Li, Yao, Zhang, Aminabadi, Awan, Rasley, and He]{rajbhandari2022deepspeed}
Samyam Rajbhandari, Conglong Li, Zhewei Yao, Minjia Zhang, Reza~Yazdani Aminabadi, Ammar~Ahmad Awan, Jeff Rasley, and Yuxiong He.
\newblock Deepspeed-moe: Advancing mixture-of-experts inference and training to power next-generation ai scale.
\newblock In \emph{International Conference on Machine Learning}, pages 18332--18346. PMLR, 2022.

\bibitem[Aminabadi et~al.(2022)Aminabadi, Rajbhandari, Awan, Li, Li, Zheng, Ruwase, Smith, Zhang, Rasley, et~al.]{aminabadi2022deepspeed}
Reza~Yazdani Aminabadi, Samyam Rajbhandari, Ammar~Ahmad Awan, Cheng Li, Du~Li, Elton Zheng, Olatunji Ruwase, Shaden Smith, Minjia Zhang, Jeff Rasley, et~al.
\newblock Deepspeed-inference: enabling efficient inference of transformer models at unprecedented scale.
\newblock In \emph{SC22: International Conference for High Performance Computing, Networking, Storage and Analysis}, pages 1--15. IEEE, 2022.

\bibitem[Li et~al.(2023)Li, Zhang, Yadav, Sung, Cheng, Bansal, and Chen]{li2023merge}
Pingzhi Li, Zhenyu Zhang, Prateek Yadav, Yi-Lin Sung, Yu~Cheng, Mohit Bansal, and Tianlong Chen.
\newblock Merge, then compress: Demystify efficient smoe with hints from its routing policy.
\newblock \emph{arXiv preprint arXiv:2310.01334}, 2023.

\bibitem[Eliseev and Mazur(2023)]{eliseev2023fast}
Artyom Eliseev and Denis Mazur.
\newblock Fast inference of mixture-of-experts language models with offloading.
\newblock \emph{arXiv preprint arXiv:2312.17238}, 2023.

\bibitem[Kim et~al.(2023)Kim, Fahim, and Awadalla]{kim2023mixture}
Young~Jin Kim, Raffy Fahim, and Hany~Hassan Awadalla.
\newblock Mixture of quantized experts (moqe): Complementary effect of low-bit quantization and robustness.
\newblock \emph{arXiv preprint arXiv:2310.02410}, 2023.

\bibitem[Shen et~al.(2023{\natexlab{a}})Shen, Hou, Zhou, Du, Longpre, Wei, Chung, Zoph, Fedus, Chen, et~al.]{shen2023mixture}
Sheng Shen, Le~Hou, Yanqi Zhou, Nan Du, Shayne Longpre, Jason Wei, Hyung~Won Chung, Barret Zoph, William Fedus, Xinyun Chen, et~al.
\newblock Mixture-of-experts meets instruction tuning: A winning combination for large language models.
\newblock \emph{arXiv preprint arXiv:2305.14705}, 2023{\natexlab{a}}.

\bibitem[Zadouri et~al.(2023)Zadouri, {\"U}st{\"u}n, Ahmadian, Ermi{\c{s}}, Locatelli, and Hooker]{zadouri2023pushing}
Ted Zadouri, Ahmet {\"U}st{\"u}n, Arash Ahmadian, Beyza Ermi{\c{s}}, Acyr Locatelli, and Sara Hooker.
\newblock Pushing mixture of experts to the limit: Extremely parameter efficient moe for instruction tuning.
\newblock \emph{arXiv preprint arXiv:2309.05444}, 2023.

\bibitem[Lewis et~al.(2021)Lewis, Bhosale, Dettmers, Goyal, and Zettlemoyer]{lewis2021base}
Mike Lewis, Shruti Bhosale, Tim Dettmers, Naman Goyal, and Luke Zettlemoyer.
\newblock Base layers: Simplifying training of large, sparse models.
\newblock In \emph{International Conference on Machine Learning}, pages 6265--6274. PMLR, 2021.

\bibitem[Roller et~al.(2021)Roller, Sukhbaatar, Weston, et~al.]{roller2021hash}
Stephen Roller, Sainbayar Sukhbaatar, Jason Weston, et~al.
\newblock Hash layers for large sparse models.
\newblock \emph{Advances in Neural Information Processing Systems}, 34:\penalty0 17555--17566, 2021.

\bibitem[Gururangan et~al.(2021)Gururangan, Lewis, Holtzman, Smith, and Zettlemoyer]{gururangan2021demix}
Suchin Gururangan, Mike Lewis, Ari Holtzman, Noah~A Smith, and Luke Zettlemoyer.
\newblock Demix layers: Disentangling domains for modular language modeling.
\newblock \emph{arXiv preprint arXiv:2108.05036}, 2021.

\bibitem[Dikkala et~al.(2023)Dikkala, Ghosh, Meka, Panigrahy, Vyas, and Wang]{dikkala2023benefits}
Nishanth Dikkala, Nikhil Ghosh, Raghu Meka, Rina Panigrahy, Nikhil Vyas, and Xin Wang.
\newblock On the benefits of learning to route in mixture-of-experts models.
\newblock In \emph{Proceedings of the 2023 Conference on Empirical Methods in Natural Language Processing}, pages 9376--9396, 2023.

\bibitem[Sukhbaatar et~al.(2024)Sukhbaatar, Golovneva, Sharma, Xu, Lin, Rozi{\`e}re, Kahn, Li, Yih, Weston, et~al.]{sukhbaatar2024branch}
Sainbayar Sukhbaatar, Olga Golovneva, Vasu Sharma, Hu~Xu, Xi~Victoria Lin, Baptiste Rozi{\`e}re, Jacob Kahn, Daniel Li, Wen-tau Yih, Jason Weston, et~al.
\newblock Branch-train-mix: Mixing expert llms into a mixture-of-experts llm.
\newblock \emph{arXiv preprint arXiv:2403.07816}, 2024.

\bibitem[Lepikhin et~al.(2020)Lepikhin, Lee, Xu, Chen, Firat, Huang, Krikun, Shazeer, and Chen]{lepikhin2020gshard}
Dmitry Lepikhin, HyoukJoong Lee, Yuanzhong Xu, Dehao Chen, Orhan Firat, Yanping Huang, Maxim Krikun, Noam Shazeer, and Zhifeng Chen.
\newblock Gshard: Scaling giant models with conditional computation and automatic sharding.
\newblock \emph{arXiv preprint arXiv:2006.16668}, 2020.

\bibitem[Zoph et~al.(2022)Zoph, Bello, Kumar, Du, Huang, Dean, Shazeer, and Fedus]{zoph2022st}
Barret Zoph, Irwan Bello, Sameer Kumar, Nan Du, Yanping Huang, Jeff Dean, Noam Shazeer, and William Fedus.
\newblock St-moe: Designing stable and transferable sparse expert models.
\newblock \emph{arXiv preprint arXiv:2202.08906}, 2022.

\bibitem[Wei et~al.(2021)Wei, Bosma, Zhao, Guu, Yu, Lester, Du, Dai, and Le]{wei2021finetuned}
Jason Wei, Maarten Bosma, Vincent~Y Zhao, Kelvin Guu, Adams~Wei Yu, Brian Lester, Nan Du, Andrew~M Dai, and Quoc~V Le.
\newblock Finetuned language models are zero-shot learners.
\newblock \emph{arXiv preprint arXiv:2109.01652}, 2021.

\bibitem[Wortsman et~al.(2022)Wortsman, Ilharco, Gadre, Roelofs, Gontijo-Lopes, Morcos, Namkoong, Farhadi, Carmon, Kornblith, et~al.]{wortsman2022model}
Mitchell Wortsman, Gabriel Ilharco, Samir~Ya Gadre, Rebecca Roelofs, Raphael Gontijo-Lopes, Ari~S Morcos, Hongseok Namkoong, Ali Farhadi, Yair Carmon, Simon Kornblith, et~al.
\newblock Model soups: averaging weights of multiple fine-tuned models improves accuracy without increasing inference time.
\newblock In \emph{International Conference on Machine Learning}, pages 23965--23998. PMLR, 2022.

\bibitem[Ainsworth et~al.(2022)Ainsworth, Hayase, and Srinivasa]{ainsworth2022git}
Samuel~K Ainsworth, Jonathan Hayase, and Siddhartha Srinivasa.
\newblock Git re-basin: Merging models modulo permutation symmetries.
\newblock \emph{arXiv preprint arXiv:2209.04836}, 2022.

\bibitem[Yadav et~al.(2024)Yadav, Tam, Choshen, Raffel, and Bansal]{yadav2024ties}
Prateek Yadav, Derek Tam, Leshem Choshen, Colin~A Raffel, and Mohit Bansal.
\newblock Ties-merging: Resolving interference when merging models.
\newblock \emph{Advances in Neural Information Processing Systems}, 36, 2024.

\bibitem[Shen et~al.(2023{\natexlab{b}})Shen, Zhang, Cao, Tan, Chen, and Gan]{shen2023moduleformer}
Yikang Shen, Zheyu Zhang, Tianyou Cao, Shawn Tan, Zhenfang Chen, and Chuang Gan.
\newblock Moduleformer: Learning modular large language models from uncurated data.
\newblock \emph{arXiv preprint arXiv:2306.04640}, 2023{\natexlab{b}}.

\bibitem[Muqeeth et~al.(2023)Muqeeth, Liu, and Raffel]{muqeeth2023soft}
Mohammed Muqeeth, Haokun Liu, and Colin Raffel.
\newblock Soft merging of experts with adaptive routing.
\newblock \emph{arXiv preprint arXiv:2306.03745}, 2023.

\bibitem[Ng et~al.(2001)Ng, Jordan, and Weiss]{ng2001spectral}
Andrew Ng, Michael Jordan, and Yair Weiss.
\newblock On spectral clustering: Analysis and an algorithm.
\newblock \emph{Advances in neural information processing systems}, 14, 2001.

\bibitem[Chi et~al.(2022)Chi, Dong, Huang, Dai, Ma, Patra, Singhal, Bajaj, Song, Mao, et~al.]{chi2022representation}
Zewen Chi, Li~Dong, Shaohan Huang, Damai Dai, Shuming Ma, Barun Patra, Saksham Singhal, Payal Bajaj, Xia Song, Xian-Ling Mao, et~al.
\newblock On the representation collapse of sparse mixture of experts.
\newblock \emph{Advances in Neural Information Processing Systems}, 35:\penalty0 34600--34613, 2022.

\bibitem[Zhou et~al.(2022)Zhou, Lei, Liu, Du, Huang, Zhao, Dai, Le, Laudon, et~al.]{zhou2022mixture}
Yanqi Zhou, Tao Lei, Hanxiao Liu, Nan Du, Yanping Huang, Vincent Zhao, Andrew~M Dai, Quoc~V Le, James Laudon, et~al.
\newblock Mixture-of-experts with expert choice routing.
\newblock \emph{Advances in Neural Information Processing Systems}, 35:\penalty0 7103--7114, 2022.

\bibitem[Dai et~al.(2022)Dai, Dong, Ma, Zheng, Sui, Chang, and Wei]{dai2022stablemoe}
Damai Dai, Li~Dong, Shuming Ma, Bo~Zheng, Zhifang Sui, Baobao Chang, and Furu Wei.
\newblock Stablemoe: Stable routing strategy for mixture of experts.
\newblock \emph{arXiv preprint arXiv:2204.08396}, 2022.

\bibitem[Fedus et~al.(2022{\natexlab{b}})Fedus, Dean, and Zoph]{fedus2022review}
William Fedus, Jeff Dean, and Barret Zoph.
\newblock A review of sparse expert models in deep learning.
\newblock \emph{arXiv preprint arXiv:2209.01667}, 2022{\natexlab{b}}.

\bibitem[Gale et~al.(2023)Gale, Narayanan, Young, and Zaharia]{gale2023megablocks}
Trevor Gale, Deepak Narayanan, Cliff Young, and Matei Zaharia.
\newblock Megablocks: Efficient sparse training with mixture-of-experts.
\newblock \emph{Proceedings of Machine Learning and Systems}, 5, 2023.

\bibitem[Li et~al.(2022)Li, Gururangan, Dettmers, Lewis, Althoff, Smith, and Zettlemoyer]{li2022branch}
Margaret Li, Suchin Gururangan, Tim Dettmers, Mike Lewis, Tim Althoff, Noah~A Smith, and Luke Zettlemoyer.
\newblock Branch-train-merge: Embarrassingly parallel training of expert language models.
\newblock \emph{arXiv preprint arXiv:2208.03306}, 2022.

\bibitem[Kudugunta et~al.(2021)Kudugunta, Huang, Bapna, Krikun, Lepikhin, Luong, and Firat]{kudugunta2021beyond}
Sneha Kudugunta, Yanping Huang, Ankur Bapna, Maxim Krikun, Dmitry Lepikhin, Minh-Thang Luong, and Orhan Firat.
\newblock Beyond distillation: Task-level mixture-of-experts for efficient inference.
\newblock \emph{arXiv preprint arXiv:2110.03742}, 2021.

\bibitem[Jaszczur et~al.(2021)Jaszczur, Chowdhery, Mohiuddin, Kaiser, Gajewski, Michalewski, and Kanerva]{jaszczur2021sparse}
Sebastian Jaszczur, Aakanksha Chowdhery, Afroz Mohiuddin, Lukasz Kaiser, Wojciech Gajewski, Henryk Michalewski, and Jonni Kanerva.
\newblock Sparse is enough in scaling transformers.
\newblock \emph{Advances in Neural Information Processing Systems}, 34:\penalty0 9895--9907, 2021.

\bibitem[Kim et~al.(2021)Kim, Awan, Muzio, Salinas, Lu, Hendy, Rajbhandari, He, and Awadalla]{kim2021scalable}
Young~Jin Kim, Ammar~Ahmad Awan, Alexandre Muzio, Andres Felipe~Cruz Salinas, Liyang Lu, Amr Hendy, Samyam Rajbhandari, Yuxiong He, and Hany~Hassan Awadalla.
\newblock Scalable and efficient moe training for multitask multilingual models.
\newblock \emph{arXiv preprint arXiv:2109.10465}, 2021.

\bibitem[Ni et~al.(2023)Ni, Iyer, Radev, Stoyanov, Yih, Wang, and Lin]{ni2023lever}
Ansong Ni, Srini Iyer, Dragomir Radev, Veselin Stoyanov, Wen-tau Yih, Sida Wang, and Xi~Victoria Lin.
\newblock Lever: Learning to verify language-to-code generation with execution.
\newblock In \emph{International Conference on Machine Learning}, pages 26106--26128. PMLR, 2023.

\bibitem[Sarkar and Lausen(2023)]{sarkar2023testing}
Soumajyoti Sarkar and Leonard Lausen.
\newblock Testing the limits of unified sequence to sequence llm pretraining on diverse table data tasks.
\newblock \emph{arXiv preprint arXiv:2310.00789}, 2023.

\bibitem[Ahn et~al.(2024)Ahn, Verma, Lou, Liu, Zhang, and Yin]{ahn2024large}
Janice Ahn, Rishu Verma, Renze Lou, Di~Liu, Rui Zhang, and Wenpeng Yin.
\newblock Large language models for mathematical reasoning: Progresses and challenges.
\newblock \emph{arXiv preprint arXiv:2402.00157}, 2024.

\bibitem[Sarkar et~al.(2022)Sarkar, Lin, Sengupta, Lausen, Zha, and Mansour]{sarkar2022parameter}
Soumajyoti Sarkar, Kaixiang Lin, Sailik Sengupta, Leonard Lausen, Sheng Zha, and Saab Mansour.
\newblock Parameter and data efficient continual pre-training for robustness to dialectal variance in arabic.
\newblock \emph{arXiv preprint arXiv:2211.03966}, 2022.

\bibitem[Hwang et~al.(2023)Hwang, Cui, Xiong, Yang, Liu, Hu, Wang, Salas, Jose, Ram, et~al.]{hwang2023tutel}
Changho Hwang, Wei Cui, Yifan Xiong, Ziyue Yang, Ze~Liu, Han Hu, Zilong Wang, Rafael Salas, Jithin Jose, Prabhat Ram, et~al.
\newblock Tutel: Adaptive mixture-of-experts at scale.
\newblock \emph{Proceedings of Machine Learning and Systems}, 5, 2023.

\bibitem[Conneau et~al.(2019)Conneau, Khandelwal, Goyal, Chaudhary, Wenzek, Guzm{\'a}n, Grave, Ott, Zettlemoyer, and Stoyanov]{conneau2019unsupervised}
Alexis Conneau, Kartikay Khandelwal, Naman Goyal, Vishrav Chaudhary, Guillaume Wenzek, Francisco Guzm{\'a}n, Edouard Grave, Myle Ott, Luke Zettlemoyer, and Veselin Stoyanov.
\newblock Unsupervised cross-lingual representation learning at scale.
\newblock \emph{arXiv preprint arXiv:1911.02116}, 2019.

\bibitem[Xue et~al.(2020)Xue, Constant, Roberts, Kale, Al-Rfou, Siddhant, Barua, and Raffel]{xue2020mt5}
Linting Xue, Noah Constant, Adam Roberts, Mihir Kale, Rami Al-Rfou, Aditya Siddhant, Aditya Barua, and Colin Raffel.
\newblock mt5: A massively multilingual pre-trained text-to-text transformer.
\newblock \emph{arXiv preprint arXiv:2010.11934}, 2020.

\end{thebibliography}
\appendix

\clearpage 
\section*{Appendix: Revisiting SMoE Language Models by Evaluating Inefficiencies with Task Specific Expert Pruning}
\addcontentsline{toc}{section}{Appendix: Optimizing SMoE Pretraining by Evaluating Inefficiencies with Task-Specific Expert Pruning} 

\FloatBarrier 

\begin{figure}[!h]
\centering
\minipage{0.45\textwidth}
\centering
\includegraphics[width=6.5cm, height=4.5cm]{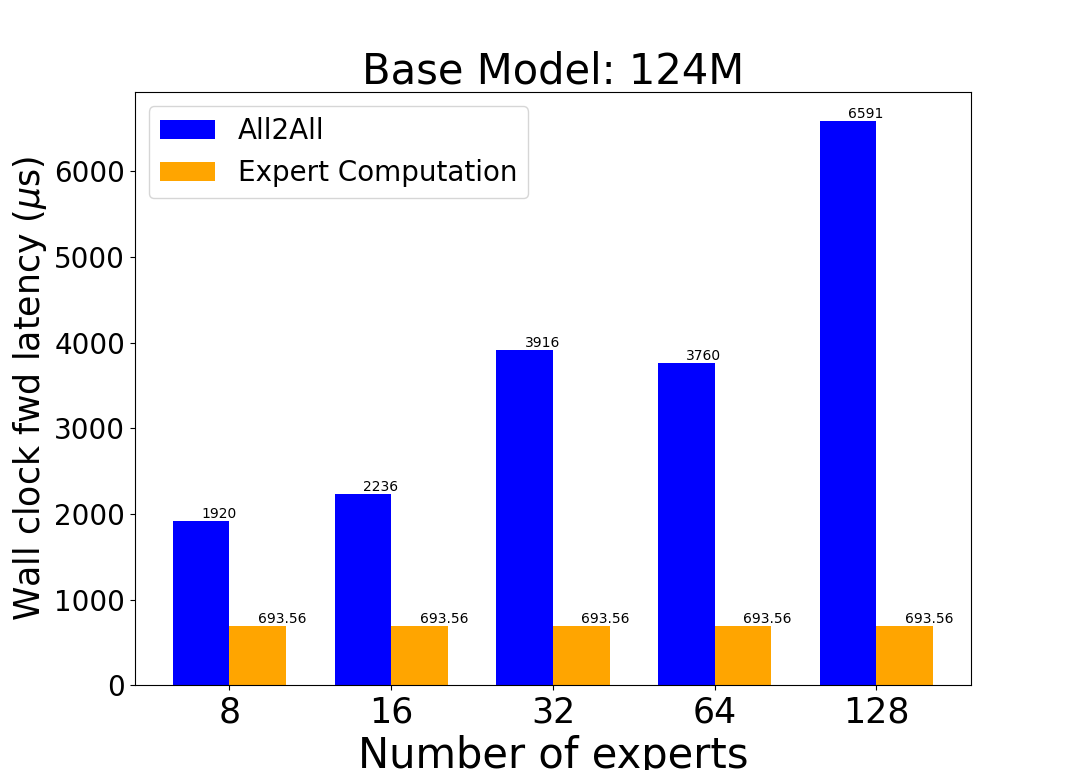}
\caption{Wall clock latencies comparing the time spent in All2All and the expert computation in one SMoE layer of a transformer block as we increase the number of experts. Expert computation time denotes the time spent in the expert FFN operations of one SMoE layer. Here, we consider a GPT2 SMoE model with backbone size of 124M parameters. Latencies are for computing the output logits of a single sequence of 512 tokens, batch size of 1 per GPU (we do not do any token generation here). Setup used with Deepspeed Zero 2 data parallelism with rank 8 and expert parallelism with rank 2.}
\label{fig:latencies}
\endminipage
\hfill
\minipage{0.45\textwidth}
\centering
\includegraphics[width=6.3cm, height=3.5cm]{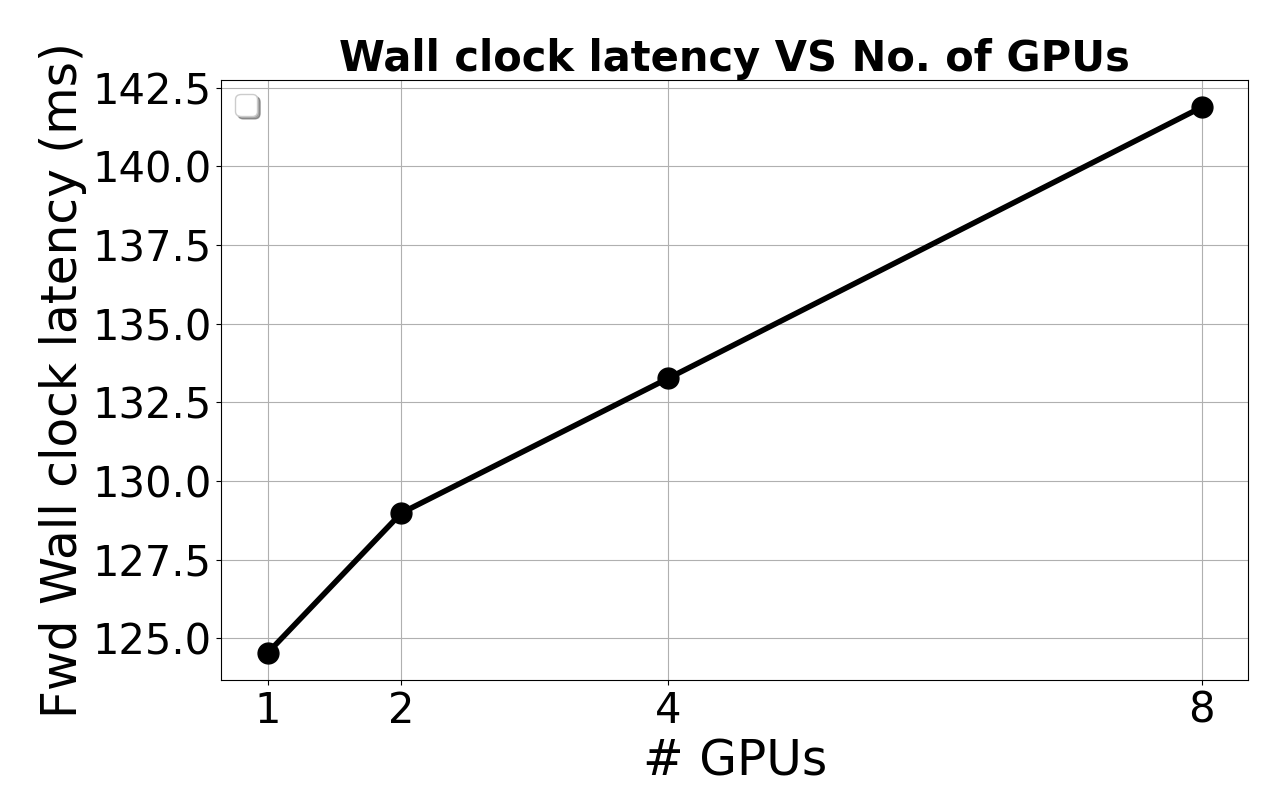}
\caption{Wall clock latencies comparing the inference time spent for a {\tt 124M+32e} SMoE model with top-1 routing as we increase the number of GPUs. Setup used with Deepspeed Zero 2 data parallelism with rank 8 and expert parallelism with rank 2, batch size of 1 per GPU and a sequence length of 256 tokens.}
\label{fig:latencies_gpu}
\endminipage
\hfill
\end{figure}
\subsection*{A1: Inference Issues with Expert Parallelism in SMoE models}
\label{sec:A1}
We start with the conventional transformer based GPT2 architecture which is the basis of all our experiments in the paper. Such a model has $N_l$ transformer layers, a hidden dimension of $d$ and context length $L$. In atransformer block, we can split the FLOPs into three groups: the dense layers in the feedforward layer (FFN), the query, key, value, and output projections (QKVO), and calculating the query-key scores and the weighted sum over the value embeddings (Attention). The FLOPs per token for the layers are calculated as follow: \textbf{FLOPs(FFN)} $=$ $N_l(24d^2)$, multi-head \textbf{FLOPs(QKVO)} $=$ $N_l(48d^2)$ and \textbf{FLOPs(Attention)} $=$ $N_l(6d(L+1))$. When the context length $L$ and $d$ are similar, the FLOPs of the FFN layer and Attention are proportionally of the same magnitude. Despite that, the FFN layer involves dense matrix multiplications, which are highly memory-intensive operations and accessing and moving data in and out of GPU memory can be a bottleneck leading to higher I/O latencies for the FFN layer. Due to this, SMoEs have focused on FFN layers to reduce inference time latencies and numerous studies in the recent past \cite{lepikhin2020gshard,fedus2022switch,du2022glam} try scalable alternatives to the dense models by reducing the computations needed for FFN. Most of the efforts in SMoE models have been to expand the FFN in a way such that a subset of the neurons in the feedforward network are activated per input through a gating mechanism. 

SMoE models, while requiring less computation over a batch of tokens compared to dense models of equivalent capacity, demand significantly more memory. To mitigate this, GShard \cite{lepikhin2020gshard} proposes expert parallelism, distributing MoE layers across multiple devices. Each device holds a subset of expert FFNs and a full copy of other parameters. Along with data parallelism where tokens are distributed across devices, for token processing by experts on different devices, an all-to-all communication collective is used to transfer tokens to the appropriate devices and back after expert processing \cite{hwang2023tutel}.

As mentioned in \cite{huang2023towards}, although theoretically SMoE models with top-1 routing execute the same number of FLOPs compared to the baseline dense models, in practice they are significantly slower. The wall clock latency difference can be attributed to four main factors: size of batched requests in terms of number of tokens, number of GPUs needed to host the model, number of experts per MoE layer and the computation costs of the expert (each FFN) itself. To understand why there is a tradeoff among these factors, we first show the costs of expert computation and the \textit{All2all} communication without considering any expert parallelism. In the case of top-K routing, the costs of the softmax operation from the routing operation has the least latency. What follows is that the latency cost is now a function of two factors: the time it takes for expert computation and the time for token to expert $All2all$ communication. From Figure~\ref{fig:latencies}, we observe that All2All dominates the expert computation in the overall inference time profile with 8 GPUs for a 124M GPT2 model as we increase the number of experts. Figure~\ref{fig:latencies_gpu} show that for a 124M model with 32 experts per MoE layer, where each alternate FFN layer in the dense 124M  model is replaced by an MoE layer of 32 experts (we denote this configuration as {\tt 124M+32e}),  the latency increases almost linearly as we scale the number of GPUs in an effort to increase throughput. This raises the question we address in our paper: what is the advantage of having more experts in an SMoE model when the latencies grow linearly with more GPUs and experts?  Is there any scope and mechanism to take advantage of the larger SMoE models with task specific pruning that can also outperform SMoE models with fewer experts, while mitigating latency overheads?


\subsection*{A2: Pretraining data}
We consider the following datasets for our study. We randomly shuffle the data between the two datasets and train them based on the hyper-parameters described next. Since we use linear scheduler for the learning rate decay, we do not consider the total number of tokens/steps to decide the decay and instead take the checkpoints after training a certain number of tokens.

\begin{itemize}
    \item \textbf{CC100}: It is a large-scale, multilingual corpus designed for training language models and other NLP tasks \cite{conneau2019unsupervised}. It is sourced from the Common Crawl, a web archive that captures vast amounts of text data from the internet. We utilize the English portion of this dataset.
    \item \textbf{mC4}: It is an extensive multilingual dataset also derived from the Common Crawl, covering 108 languages \cite{xue2020mt5}. It is a subset of the Colossal Clean Crawled Corpus (C4), which itself is a cleaned version of the Common Crawl.  As before, we utilize the English portion of this dataset.
\end{itemize}

\begin{table}[t!]
\centering
\caption{Terminologies used for MoE layers adopted from \cite{zoph2022st}}
\begin{tabular}{|l|p{10cm}|}
\hline
\textbf{Terminology} & \textbf{Definition} \\ \hline
\textbf{Expert} & An independently-learned neural network with unique weights. \\ \hline
\textbf{Router} & A network that computes the probability of each token getting sent to each expert. \\ \hline
\textbf{Top-n Routing} & Routing algorithm where each token is routed to $n$ experts. \\ \hline
\textbf{Load Balancing Loss} & An auxiliary (aux) loss to encourage each group of tokens to evenly distribute across experts. \\ \hline
\textbf{Capacity Factor (CF)} & Each expert can only process up to a fixed number of tokens, which is often set by evenly dividing across experts, $\frac{\text{tokens}}{\text{experts}}$. The capacity factor can expand or contract this amount to CF $\cdot \frac{\text{tokens}}{\text{experts}}$. \\ \hline
\textbf{FFN} & Acronym of Feed Forward Network (FFN) layer of Transformer consisting of linear, activation, linear. \\ \hline
\end{tabular}
\end{table}

\subsection*{A3: Pretraining Models and Hyper-parameters}
The model architectures of the starting dense models in our work are shown in Table~\ref{tab:arch}. We use the same byte pair encoding (BPE) tokenizer used in GPT2 in our training pipeline.
\begin{table}[!h]
\centering
\begin{tabular}{p{1.5cm}|p{1.3cm}|p{1cm}|p{1cm}|p{1.5cm}}
\toprule
\textbf{Model} & $d_{model}$ & $d_{ff}$ & $\#$ heads & $\#$ layers   \\ \midrule
\textbf{124M}  & 1024 & 4096 & 12 & 12  \\ \midrule
\textbf{354M}  & 1024 & 4096 & 16 & 24  \\ \midrule
\end{tabular}
\caption{Model architectures of the dense backbone GPT2 models.}
\label{tab:arch}
\end{table}
For the SMoE implementation, we used the top-$\mathbf{K}$ layer class used in Deepspeed-MoE \footnote{https://www.deepspeed.ai/tutorials/mixture-of-experts-nlg/} in our implementations.  We use an input sequence length of 1024 tokens for all our pretraining runs.  Note that we pack multiple sequences into the same input by using padding tokens when the inputs sequences do not fill 1024 tokens.  For all the training runs, we set the peak learning rate to $5e^{-5}$ and Table~\ref{tab:train_hypers} lists the hyper-parameters used in the pretraining stage of the SMoE models. We reiterate that for all our pre-training runs, we start with randomized initialization of the model families without changing the architectures.  We train all the checkpoints for a total of 40B tokens on the mix of CC100 and mC4 English data described above. For pretraining each of the models {\tt 354M}, {\tt 354M+8e}, {\tt 354M+32e}, {\tt 354M+64e}, {\tt 354M+128e} to 40B tokens, we utilize 4 p4de nodes\footnote{https://aws.amazon.com/ec2/instance-types/p4/}, each having 8 A100 GPUs with 80 GB memory. 
 
We refer the reader to \cite{zoph2022st} on more details of the hyper-parameters used for the top-k routing based implementation of MoE layers.

\begin{table}[!t]
\caption{Hyper-parameters used in pretraining the SMoE models.}
\centering
\begin{tabular}{p{6cm}|p{2cm}}
\toprule
\textbf{Hyper-parameter} & \textbf{Value}   \\ \midrule
Train Capacity factor  & 1.2   \\ \midrule
Evaluation Capacity factor  & 1  \\ \midrule
Minimum Expert Capacity  & 4   \\ \midrule
Drop Tokens & True   \\ \midrule
Load Balancing loss coeff. $\alpha$ & 0.01   \\ \midrule
Batch size (combined across GPUs) & 512   \\ \midrule
Learning Rate Scheduler & Linear \\ \midrule
Input Sequence Length & 1024  \\ \midrule
Peak Learning Rate & 5e-5 \\ \midrule
gradient\_clip\_val & 1.0 \\ \midrule
Warmup steps & 1000 \\ \midrule
Weight decay & 0.1 \\ \midrule
Optimizer & AdamW \\ \midrule
$\epsilon$, $\beta_1$, $\beta_2$ & 1e-8, 0.9, 0.95\\ \bottomrule
\end{tabular}
\label{tab:train_hypers}
\end{table}

\begin{figure}[!t]
\centering
\centering
\includegraphics[width=6.5cm, height=4.5cm]{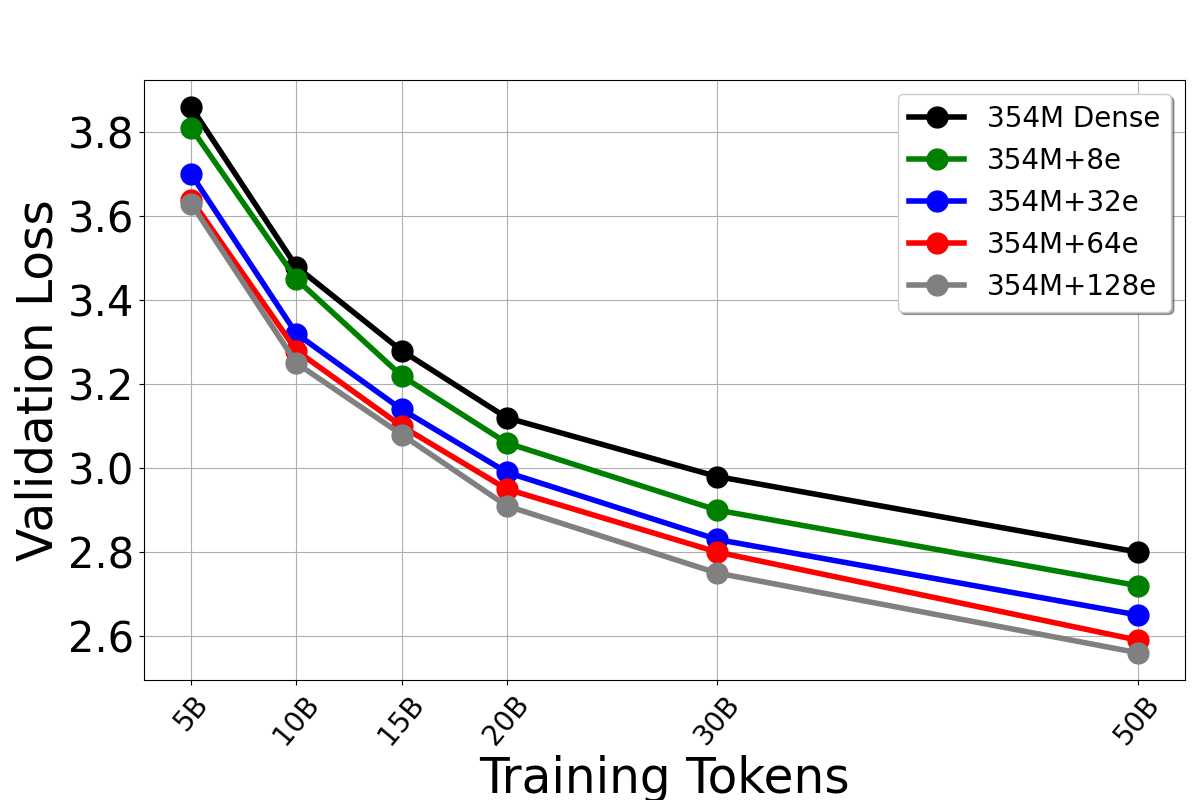}
\caption{Training Loss curves.}
\label{fig:lossc}
\end{figure}

We plot the training loss curves in Figure~\ref{fig:lossc} and we observe that even in the overtrained regime for the {\tt 354M+8e} (1.3B) model, for the 30B tokens (against the Chinchilla [2] optimal of 20B tokens), the validation loss is higher than the {\tt 354M+128e} (13B) trained for 50B tokens (which is undertrained as per Chinchilla). This shows that the smaller models need to be trained for a lot more tokens than optimal as compared to the larger number of experts and in the finite data regime (same pretraining FLOPs irrespective of total number of experts), SMoEs with more experts have an advantage.

\subsection*{A4: Finetuning Data and Hyper-parameters}
We list the hyper-parameters for the single-task finetuning case in Table~\ref{tab:eval_hypers}. We use an input sequence length of 256 for the finetuning and use the same top-1 routing with the same load balancing loss and the balancing coefficients as used in pretraining. We do not use example packing during  this stage of the training and evaluation. For the subset of the SuperGLUE tasks \footnote{https://super.gluebenchmark.com/tasks} on which we evaluate the models, we use promptsource \url{https://github.com/bigscience-workshop/promptsource} to create the inputs and the outputs. 

For finetuning the models with baselines mentioned in Section~\ref{sec:baselines}, we use the same hyper-parameters as those used on our models.

\begin{table}[!h]
\caption{Hyper-parameters used in finetuning the SMoE models.}
\centering
\begin{tabular}{p{6cm}|p{2cm}}
\toprule
\textbf{Hyper-parameter} & \textbf{Value}   \\ \midrule
Train Capacity factor  & 1.2   \\ \midrule
Evaluation Capacity factor  & 1  \\ \midrule
Minimum Expert Capacity  & 4   \\ \midrule
Drop Tokens & True   \\ \midrule
Load Balancing loss coeff. $\alpha$ & 0.01   \\ \midrule
Batch size (combined across GPUs) & 64   \\ \midrule
Input Sequence Length & 128  \\ \midrule
Peak Learning Rate & 1e-6 \\ \midrule
Learning Rate Scheduler & Linear \\ \midrule
gradient\_clip\_val & 1.0 \\ \midrule
Warmup steps & 500 \\ \midrule
Number of Epochs & 50 \\ \midrule
Weight decay & 0.001 \\ \midrule
Optimizer & AdamW \\ \midrule
$\epsilon$, $\beta_1$, $\beta_2$ & 1e-8, 0.9, 0.95\\ \bottomrule
\end{tabular}
\label{tab:eval_hypers}
\end{table}

 \pagebreak
\subsection*{A5. Examples of Inputs and Outputs in the Finetuning and Evaluation data}

\begin{table}[h!]
\centering
\begin{tabular}{|l|p{12cm}|}
\hline
\textbf{Task} & \textbf{Example} \\ \hline
\textbf{COPA} & \textbf{Input}: The woman tolerated her friend's difficult behavior.\textbackslash n \textbackslash n Select the most plausible  cause: \textbackslash n- The woman knew her friend was going through a hard time.\textbackslash n-  The woman felt that her friend took advantage of her kindness. \newline \textbf{Output}: The woman knew her friend was going through a hard time. \\ \hline
\textbf{RTE} & \textbf{Input}: Given that Dana Reeve, the widow of the actor Christopher Reeve, has died of lung cancer at age 44, according to the Christopher Reeve Foundation. Does it follow that Christopher Reeve had an accident. Yes or no? \newline \textbf{Output}: No. \\ \hline
\textbf{COPA} & \textbf{Input}: The man turned on the faucet.\textbackslash n  \textbackslash n Select the most plausible  effect: \textbackslash n- The toilet filled with water.\textbackslash n- Water flowed from the spout. \newline \textbf{Output}: The woman knew her friend was going through a hard time. \\ \hline
\textbf{WSC} & \textbf{Input}: Bernard , who had not told the government official that he was less than 21 when he filed for a homestead claim, did not consider that he had done anything dishonest. Still, anyone who knew that he was 19 years old could take his claim away from him . Is the coreference between anyone and him : True OR False? \newline \textbf{Output}: False. \\ \hline
\textbf{CB} & \textbf{Input}: Given that Valence the void-brain, Valence the virtuous valet. Why couldn't the figger choose his own portion of titanic anatomy to shaft? Did he think he was helping? Does it follow that Valence was helping Yes, no, or maybe? \newline \textbf{Output}: No. \\ \hline
\textbf{FLAN} & \textbf{Input}: Generate a correctly punctuated version of the following text: MeetAlpakkaTweed from Du StoreAlpakka a new and exciting yarn from the Norwegian yarn manufacturer. \newline \textbf{Output}: MeetAlpakkaTweed from Du StoreAlpakka- a new and exciting yarn from the Norwegian yarn manufacturer! \\ \hline
\end{tabular}
\end{table}

\end{document}